\documentclass[format=manuscript]{acmart}

\usepackage{graphicx} 
\usepackage{booktabs}
\usepackage{mathtools} 
\usepackage{listings}
\usepackage{longtable}
\usepackage{makecell}
\usepackage{multirow}
\usepackage{url}
\usepackage{hyperref}
\newtheorem{assumption}{Assumption}
\newtheorem{problem}{Problem}
\usepackage{changepage}           
\usepackage{seqsplit, adjustbox}
\usepackage{float}
\usepackage{pgfplots}
\usepackage{pgfplotstable}
\pgfplotsset{compat=1.18}
\usepackage{subcaption} 
  
\usepackage{CJKutf8} 

\usepackage{appendix}
\usepackage{ragged2e}   
\usepackage{array}      
\usepackage{tabularx}
 
\newcolumntype{C}{>{\RaggedRight\arraybackslash}p{0.5cm}}

\newcolumntype{Y}{>{\RaggedRight\arraybackslash\sloppy\hyphenpenalty=10000\exhyphenpenalty=10000}p{5cm}}

\title{A Generative Grammar Underlying the Voynich Manuscript, the Pastiche Hypothesis: Evidence from Large Language Models}

\author{Nicolas Turenne}
\email{nicolas.turenne@ird.fr}
\affiliation{%
  \institution{INRAE, MathNum ;
     IRD, Sorbonne Université, UMMISCO}
  \city{Paris}
  \country{France}
}

\begin{document}

\begin{abstract}
Background:
The Voynich Manuscript is a fifteenth-century codex written in an unknown script whose content remains undeciphered. Previous studies suggest that its statistical properties resemble those of natural languages, while its illustrations—primarily plants—recall medieval herbals.

Methods:
We present a multidisciplinary analysis combining probabilistic modeling, phonetic decomposition, rare-event detection, and multimodal image analysis, based on a newly transliterated corpus. Word- and letter-level distributions are modeled using position-dependent probabilistic grammars, while phonetic patterns are compared across Indo-European, Semitic, and Asian languages. Image–text alignment methods based on large language models are applied to identify potential botanical correspondences.

Results:
The results indicate that Voynich symbols behave as letters rather than syllabic units, while word-length distributions resemble syllabic structures. Phonetic analyses show closer alignment with consonant-heavy languages such as Hebrew or Arabic than with Indo-European languages. Probabilistic modeling reproduces Zipf-like distributions and reveals extremely low probabilities for repeated initial-letter sequences, indicating a structured imitation of natural language. Image analysis suggests strong correspondences between Voynich plant illustrations and those found in Pseudo-Apuleius herbals from the Mediterranean tradition, consistent with an imitation of medieval medicinal books.

Perspectives:
These findings support the hypothesis that the Voynich Manuscript follows a structured generative system combining linguistic regularities and herbal knowledge, and demonstrate the value of integrating probabilistic and AI-assisted approaches in the analysis of historical manuscripts.
\end{abstract}



\flushbottom
\maketitle

\thispagestyle{empty}


\section*{Introduction}
The Voynich Manuscript, discovered in 1912 and written in an unknown script by an unidentified author, has captivated the attention of linguists and computer scientists for several decades. Early research into the exploration of textual corpora and the analysis of written data emerged through statistical studies of texts—such as word-length frequencies, lexical usage, and sentence patterns—developed for the purpose of authorship attribution. One of the earliest contributions in this domain was made by the physicist Thomas Mendenhall (\cite{Mendenhall1887}). These pioneering approaches continue to inform contemporary research in text analysis and stylometry (\cite{Savoy2020}). Modern statistical text analysis originates in the modeling and clustering of textual units and their cohesion \cite{Benzecri1981}\cite{Halliday1976}. These approaches have since been extended to keyword extraction (\cite{Lebart1998},\cite{Firoozeh2020}), the analysis of social media data \cite{Dang2018, Turenne2023YouTube}, to the construction of embedding spaces for text classification and multilingual processing \cite{Turenne2020,Turenne2022}, and to human–machine dialogue systems based on large language models \cite{AionX2025}. Bigrams and co-occurrences of words or letters and their entropy (i.e., n-grams; \cite{Kim2010,Schurmann1996,Holzinger2014}) provide a means of encoding information compactness, in a manner analogous to Shannon’s information theory \cite{Shannon1949}.
The analysis of the Voynich Manuscript falls within this framework, as it entails the examination of a work of unknown authorship and the interpretation of its features through statistical modeling. In the contemporary context of large language models and digital humanities, such approaches aim to shed light on the manuscript’s structure, content, and possible origins. \\
Prior to the manuscript’s dating, the Voynich Manuscript was frequently claimed to be a hoax (\cite{Rugg01012004}), a position that persisted even after its official radiocarbon dating (\cite{RuggTaylor2017}). However, random or purely algorithmic models fail to reproduce the manuscript’s multi-level structural consistency—particularly its lexical cohesion, which is characteristic of ritual and herbal registers—thereby lending support to non-hoax hypotheses (\cite{Landini2001}). 
Statistical analyses by \cite{MontemurroZanette2013} further suggest that the text exhibits properties consistent with those of a natural language. Building on this line of inquiry, \cite{Altmann2016}\cite{Amancio2013} developed a statistical framework for analyzing unknown texts and assessing their compatibility with natural languages, employing metrics based on word statistics, network topology, and textual intermittency.
They validate their approach using known texts, demonstrating that the proposed metrics can reliably distinguish meaningful texts from shuffled or random sequences, and that certain measurements are more sensitive to syntactic structure than to semantic content. 
When applying this framework to the undeciphered Voynich Manuscript, they find it to be largely compatible with natural languages and identify candidate keywords that may support future efforts at decipherment.
The manuscript’s enigmatic nature has inspired a wide range of hypotheses, from deliberate hoax (\cite{Rugg01012004}) and cryptographic experimentation to proposed connections with languages as diverse as
Arabic, Hebrew, Nahuatl, or a putative proto-Romance (\cite{Bax2014Report}\cite{Crowe2022Voynich}\cite{HANNIG2020}). More recently, \cite{Gibbs2017} proposed a partial translation into Latin, focusing on newly identified glyphs interpreted as referring to herbal concepts such as folio or aromatic.\\
Despite these numerous attempts to translate the Voynich Manuscript, scholarly articles continue to emphasize its persistent strangeness and resistance to interpretation (\cite{Livingstone2016}). 
By contrast, \cite{GaskellBowern2022} asked volunteers to generate meaningless texts and compared the results to the Voynich Manuscript. 
Although they reported certain similarities, their study did not provide conclusive proof and nevertheless claimed that the manuscript is gibberish.
Regarding the visual aspects of the manuscript, \cite{TuckerTalbert2014} and \cite{JanickTucker2018} interpreted certain plant illustrations (notably folio 86v) as originating from Central Mexico. \\
Our study benefits from the fact that, since 2022, new avenues of investigation have become possible.
First, a new corpus has become available through the work of Zandbergen (\cite{zandbergen2022transliteration}). Second, large language models are now publicly accessible and integrated into information-search interfaces, most notably ChatGPT (currently version 5.2; \cite{AionX2025}).
The development of large-scale models began with AlexNet, a deep convolutional neural network introduced by Krizhevsky et al. (\cite{Krizhevsky2012}). This model dramatically reduced the top-5 error rate on the ImageNet dataset from approximately 26 \%—the previous state of the art—to about 15 \%, marking a turning point in large-scale visual recognition.
This work demonstrated the effectiveness of deep convolutional networks trained on large-scale datasets using GPUs, thereby catalyzing the modern deep learning revolution. 
Subsequently, \cite{Vaswani2017} introduced the Transformer architecture, which has since become the foundational framework for large language models, including ChatGPT developed by OpenAI and Gemini developed by Google (\cite{Radford2018}\cite{PalSankarasubbu2024}\cite{AionX2025}).\\

To investigate the nature of the Voynich Manuscript, we combine probabilistic string modeling, large language model (LLM)–based image decoding, LLM-based phonetic modeling, and LLM-based analysis of frequent letters to generate and evaluate multiple hypotheses. The key aspects of our study are as follows:

\begin{itemize}
  \item \textbf{Contribution to the resolution of the Voynich mystery through the pastiche hypothesis.}  
  The results support the hypothesis that the Voynich text is a structured pastiche, assembled from recurring symbolic units and formulaic patterns rather than encoding a natural language or ciphered message. By demonstrating how local coherence can emerge from recombination of pre-existing elements, the study offers a parsimonious and testable explanation for the manuscript’s apparent complexity and long-standing resistance to decipherment.

  \item \textbf{Methodological rigor and reproducibility.}  
  The report applies quantitative, computational methods grounded in the scientific method, with explicitly defined datasets, metrics, and analytical procedures. This ensures that the results are reproducible and falsifiable, marking a clear departure from speculative or anecdotal approaches that have traditionally dominated Voynich studies.

  \item \textbf{Data-driven insights beyond traditional close reading.}  
  Digital analysis uncovers statistical regularities, distributional patterns, and combinatorial constraints within the text that are not readily observable through manual inspection. These findings provide empirical evidence for underlying structural principles governing the manuscript’s construction.

  \item \textbf{Interdisciplinary significance and methodological transferability.}  
  By bridging digital humanities, linguistics, and computational analysis, the report illustrates how scientific methodologies can be applied to historically enigmatic texts. The approach is transferable to other undeciphered or artificial corpora, providing a general framework for investigating textual artifacts that lie at the boundary between language, code, and design.
\end{itemize}

This article is organized as follows. 
In Section~\ref{sec:vm}, we provide an historical context overview of the Voynich Manuscript. 
Section~\ref{sec:dataset} describes the digital corpus used in our study, including statistical distribution of features. Section~\ref{sec:herbal} situates the manuscript within the broader tradition of ancient herbal medicine, highlighting textual and visual conventions that inform the interpretation of botanical illustrations. In Section~\ref{sec:symbolrep}, we present our assumption for representing unknown manuscript’s glyphs. Section~\ref{sec:phonetic} explores phonetic decomposition and compares Voynich glyph sequences to compare phoneme distribution across languages. Section~\ref{sec:frequentletter} most frequent letters across languages to analyze potential encoding schemes. 
Section~\ref{sec:imagemed} examines the hypothesis that the majority of the manuscript's illustrations correspond to real medicinal plants. Section~\ref{sec:comppicture} discusses additional pictorial elements.
Section~\ref{sec:eventdet} examines unusual initial letter sequences to test potential structural or linguistic rules. 
Finally,  Section~\ref{sec:WordGen} introduces our probabilistic string model for generating candidate words and exploring distribution law.

\section{Voynich Manuscript}
\label{sec:vm}

\subsection{Historical description}

The manuscript examined in this study is referred to as the Voynich Manuscript (hereafter VM), named after the librarian Wilfrid Michael Voynich (1865–1930), who discovered it in 1912 \cite{Walters2024} at the Villa Mondragone near Rome \cite{Wikipedia2024}. Voynich retained possession of the manuscript until his death. Since 1969, it has been housed at Yale University’s Beinecke Rare Book and Manuscript Library, where it is available for open access consultation (\cite{VoynichOriginal2026}).\\
A dating experiment conducted by the University of Arizona in 2011 revealed that the materials of VM (paper and ink) date from approximately 1420 ± 20 years, placing it in the 15th century. Greg Hodgins (\cite{ScienceDaily2011}) from the Department of Physics at the University of Arizona traveled to Yale University’s Beinecke Rare Book and Manuscript Library, where the manuscript is held, to collect four micro-samples measuring 1 mm by 6 mm. Using a mass spectrometer in his laboratory, he was able to determine the carbon-14 content of these samples, which were made from animal skin.\\
This is the only aspect of the Voynich Manuscript that can be stated with certainty: the letters it employs, its content, and its author all remain unknown.\\
The manuscript is written on parchment made from sheepskin and consists of 116 folios, corresponding to 204 PDF pages. The folios are numbered in Arabic numerals from 1 to 116, although Arabic numbers appear only for page numbering. All pages are written in an unknown alphabet and language.  

The \emph{Voynich Manuscript} can be divided into three parts based on folio and PDF pagination:

\begin{itemize}
    \item \textbf{Part 1} (pdf 111 pages): folio 1r (PDF page number 2) to folio 57r (PDF page number 114)
    \item \textbf{Part 2} (pdf 69 pages): folio 57v (PDF page number 115) to folio 102v (PDF page number 183)
    \item \textbf{Part 3} (pdf 24 pages): folio 103r (PDF page number 184) to folio 116v (pdf page number 207)
\end{itemize}

\begin{figure}[!ht]
  \centering
    \centering
    \includegraphics[width=\linewidth]{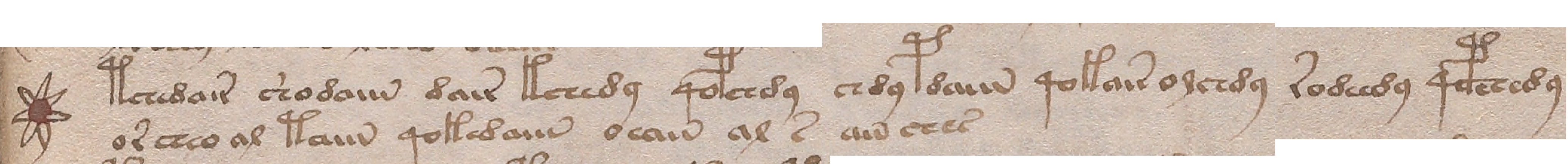}
    \caption{Example of paragraph in the VM manuscript}
  \label{fig:vm-paragraph} 
    \Description{Screenshot showing an example paragraph from the VM manuscript.}

\end{figure}

In Appendix (supplementary material) section "Voynich pages" a table lists PDF pages containing multiple folios. On 17 PDF pages, there are in fact 45 paper pages, i.e., 28 additional pages. 
Therefore, the manuscript contains a total of 232 paper pages (116 folios, recto and verso).\\
One hundred and twenty-nine plant illustrations appear in the first half of the manuscript, each accompanied by surrounding commentary. 
The last twenty pages contain no illustrations. The text can be divided into 740 paragraphs (see Figure~\ref{fig:vm-paragraph} for an example). 
There is no punctuation, except for an asterisk-like symbol that occasionally marks the beginning of a paragraph.

If one imagines a single individual composing the manuscript using an invented alphabet, the task would have required several months of sustained effort, representing a considerable investment of time and labor. 
Moreover, the work is unique in its kind and does not appear to have been intended for any specific recipient.

\subsection{Glyphs}

The \emph{Voynich Manuscript} is renowned not only because its content remains undeciphered, but also due to the use of unknown glyphs, which give the impression of a lost or unknown language (see Appendix (supplementary material page 18 ) section "Most Frequent Glyphs" for an overview of glyphs). 
The symbols themselves are inherently mysterious (see Figure~\ref{fig:vm-glyph} for an example). 
In the table provided in Appendix (supplementary material page18) section "List of 170 glyphs in Voynich manuscript", 170 glyphs are represented; however, in the dataset we used from Zandbergen (\cite{zandbergen2022transliteration}), a total of 423 glyphs have been defined for the transcription of the manuscript (\cite{zandbergen2023sta}).

\begin{figure}[ht]
  \centering
    \centering
    \includegraphics[width=0.1\linewidth]{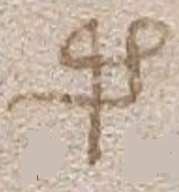}
    \caption{Example of a glyph in the VM manuscript (defined as P1 by Zandbergen see \cite{zandbergen2023sta})}
  \label{fig:vm-glyph} 
    \Description{Single glyph from the Voynich Manuscript corresponding to the symbol classified as P1 by Zandbergen.}

\end{figure}

\subsection{Established Facts}

Only a few established facts about the \emph{Voynich Manuscript} are widely accepted:  

\begin{enumerate}

    \item The manuscript is authentic and is archived at the University of Pennsylvania Library \cite{VoynichOriginal2026}.
    
    \item The manuscript resembles an artificial medieval medicinal herbarium.
    
    \item The writing dates to the 15th century (\cite{ScienceDaily2011}.
    
    \item Different sections of the manuscript appear to have been written by multiple individuals. At least three distinct hands can be identified. \cite{Currier1976} suggested that the VM could have been written in "two languages and by four to six hands," indicating two distinct contents and several different scribes.
    
    \item The book is a unique document employing this particular set of glyphs.
    
\end{enumerate}

\subsection{Special observations}

Table~\ref{tab:page_observations_en} highlights anomalous features observed in certain parts of the manuscript. The presence of a Christian cross and a crossbowman challenges the hypothesis that the work is written in Arabic or Hebrew. The observation on pdf page 156 (folio 85r) where all first paragraph symbols are identical, represents a notable constitutes a peculiarity of the manuscript.

\begin{table}[htbp]
\centering
\small
\begin{tabular}{p{0.25\textwidth} p{0.3\textwidth}}
\hline
\textbf{pdf Page / Folio} & \textbf{Observation} \\
\hline
51\ (25v) & Contains a small dragon \\
66\ (33r) & Contains two little figures \\
156  (85r) & All paragraphs start with the same symbol \\
145 (79v) & Christian cross \\
135 (73v) & Crossbowman \\
134 (73r) & Salamander \\
120 (66r) & isolated letters at beginning of each line \\
\hline
\end{tabular}
\caption{Iconographic and symbolic observations by page/folio}
\label{tab:page_observations_en}
\end{table}

\subsection{Various hypotheses}

Several hypotheses have been proposed to explain the content of the \emph{Voynich Manuscript}, including the suggestion that it is a hoax (\cite{Tiltman1967}, \cite{Dimperio1978}, \cite{Rugg01012004}). 

More recent hypotheses (\cite{HANNIG2020}, \cite{Hauer2016}) have argued that the original document was written in ancient Hebrew. However, these studies have not succeeded in producing a complete translation of the manuscript.

\subsection{Other ciphered cases}

Throughout history, thousands of ciphered texts (letter correspondence) have been produced; however, manuscripts written entirely in unknown symbols remain exceedingly rare. 
One notable example is the Rohonc Codex (Hungary, 16th century), a 448-page manuscript composed in an unidentified alphabet comprising nearly 200 distinct symbols and containing both biblical and military illustrations (\cite{Rohonczi1838}).
The codex is generally considered to be of Austro-Hungarian origin and is preserved at the University of Budapest. 
Copies have also been studied in Romania, where some scholars have explored the possibility of a connection to a putative Dacian script. 
First documented in 1838 in the library of Gusztáv Batthyány, the manuscript was later examined in Germany, where it was characterized as “indecipherable” and consisting of “meaningless writing.” \\
The Book of Angels (Liber Logaeth), attributed to John Dee and Edward Kelley in the sixteenth century, presents the Enochian language—an invented alphabet and linguistic system purportedly revealed by angels and employed for esoteric, magical, and mystical communication.\\
The Book of Soyga (Aldaraia sive Soyga vocor, sixteenth century) is a Latin magical treatise once owned and studied by John Dee (\cite{Soyga1550}). 
The text employs a mixed alphabet combining Latin characters with ciphered and symbolic forms, and encompasses magical practices, astrological and numerical tables, and complex systems of esoteric codification. 
Two manuscripts survive—Sloane MS 8 and Bodley MS 908—both of which contain extensive letter tables, including thirty-six large squares whose underlying logic remained undeciphered by Dee himself. 
In addition to incantations, the work addresses astrology, demonology, and the genealogies of angels, functioning as a structured symbolic system intended to preserve and safeguard specialized magical knowledge.

\section{Dataset}
\label{sec:dataset}

\subsection{Corpus}

\begin{figure}[htbp]
  \centering
  \includegraphics[width=\textwidth]{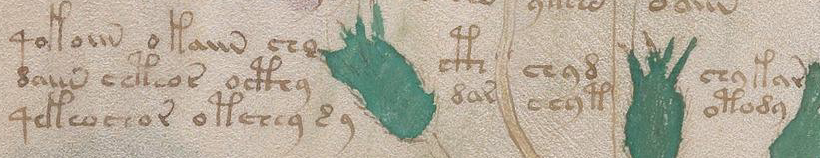}
  \caption{Voynich Manuscript, folio 29v (see \cite{VoynichOriginal2026}).}
  \label{fig:voynich29v}
    \Description{Page 29v of the Voynich Manuscript showing handwritten text and botanical-style illustrations.}

\end{figure}

In this study, we utilize the Zandbergen corpus, which has been publicly available online since 27 February 2023 (\cite{zandbergen2022transliteration, zandbergen2023sta}). 
The Zandbergen–Landini (ZL) transliteration comprises a total of 157,300 glyphs. Size on disk is 473,055 bytes.
A notable feature of the Zandbergen corpus is the inclusion of approximately 5,615 loci, corresponding to individual lines within the manuscript.\\
For reference, the manuscript comprises 116 folios, corresponding to 232 paper pages. 
Each folio includes a recto and a verso, with folios conventionally indexed as, for example, 1r and 1v for the first folio.\\
In the digitized corpus, delimiters between contiguous sequences of glyphs (i.e., spaces) are primarily represented by the dot symbol (``.'') and occur 35,525 times. 
In some cases, spaces are also indicated by a comma (``,'') with 8,125 occurrences. 
Overall, there are 43,650 delimiters separating glyph sequences, corresponding to an average of 188 per paper page. 
The most frequent glyph is A1, appearing 25,621 times; if A1 were interpreted as a delimiter, this would correspond to roughly 110 delimiters per paper page. 
Interpreting A1 as a delimiter produces entire lines without additional separators, which appears unusual. For example:

\begin{verbatim}
<f86v3.6,+Pb>     B1A3G1B2C2,A3G1.K1U2A2.A2Q1A3G1.A2Q1A3G1.K1B1A3C1.A3C1,A3G1
\end{verbatim}

Glyphs are distributed across 740 paragraphs (see Table~\ref{tab:f29v-transcription}). 
The mean number of glyphs per paragraph is 237, with the 90th percentile reaching 396 glyphs. 
These paragraphs are written over 5,617 lines (loci).
Figure~\ref{fig:voynich29v} illustrates the transcription of a short passage from the \emph{Voynich Manuscript} (\cite{VoynichOriginal2026}).

\begin{table}[ht]
\centering
\renewcommand{\arraystretch}{1.2}
\begin{tabular}{ll}
\hline
\textbf{Folio} & \textbf{Transcription} \\
\hline
f29v.10+P0 &
\texttt{<\%>D1A1Q1A1G1.A1Q1A3G1.K1A1<->[Uf:U2]<->K1A2B1<->K1A2Q1A3C1} \\
f29v.11+P0 &
\texttt{B1A3G1.K1Q2J1A1[C1:C2].A1U2A2<->B1A3C1<->K1A2Q2<->A1Q2A1B1A2} \\
f29v.12+P0 &
\texttt{D1J1Q1J1A1K1A1C1.A1Q2K1J1A2.C2,A2<\$>} \\
\hline
\end{tabular}
\caption{Transcriptions from folio f29v. The symbols \% and \$ indicate the beginning and end of a paragraph, respectively. The metadata header ``f29v.10+P0'' provides information about the folio. Each entry corresponds to three lines (i.e., loci) forming a single paragraph (\cite{VoynichZL3a2026}).}
\label{tab:f29v-transcription}
\end{table}

\subsection{Computing facilities and hardware configuration}

All computational analyses presented in this study were conducted on a dedicated workstation with the following specifications. 
The machine is equipped with a 900~GB storage disk and 32~GB of RAM. It runs on an Intel\textsuperscript{\textregistered} Core\textsuperscript{\texttrademark} Ultra 7 165U processor operating at 1.70~GHz.\\
The operating system is a 64-bit version of Windows~11~Professional. The hardware platform corresponds to an HP EliteBook 860 G11 (16-inch) PolyStudio model. 
This configuration provided sufficient computational capacity to perform the corpus processing, statistical analyses, and visualization tasks required in this study.

\subsection{Corpus preprocessing}

Minimal preprocessing was applied to the corpus in order to preserve its original structure:

\begin{itemize}

  \item Commas were replaced with dots and used 
  consistently as glyph sequence delimiters.
  
  \item The symbol sequence ``<->'' was replaced by a dot.
  
  \item Uncertainty brackets were removed and replaced with the first candidate glyph; for example, \texttt{[A1:A2]} was transformed into \texttt{A1}.
  
\end{itemize}

\subsection{Letter classification}

In Table~\ref{tab:symbol_families}, the distribution of glyphs according to their membership in symbol families is presented. 
Some symbols are grouped within the same family in the corpus due to their graphical similarity. 
Overall, the corpus identifies 423 distinct glyphs distributed across 23 families, with the X and Z families assigned a special status, leaving 21 families considered significant. \\
Of the 423 distinct glyphs, only 47 occur more than 10 times. 
Despite representing a small fraction of the total glyph inventory, these 47 frequent glyphs account for 156,703 occurrences out of 157,300, corresponding to 99.62\% of all glyph instances (see table In Appendix (supplementary material page18) section "List of 170 glyphs in Voynich manuscript").\\
In terms of symbolic families, glyphs belonging to family A dominate the dataset, representing 37.1\% of total occurrences, while family B accounts for 16.0\%. 
Taken together, families A and B comprise the majority of the corpus, totaling 53.1\% of all glyphs occurrences. 
Each of the remaining families contributes less than 7\% of total occurrences, indicating a highly skewed distribution concentrated in a small number of dominant symbolic families.\\
Hereafter, we will use the terms ``symbol,'' ``glyph,'' and ``letter'' interchangeably to refer to the same entity.

\begin{longtable}{|p{1cm}|p{1.3cm}|p{6cm}|l|}
\caption{Symbol families with structural descriptions and representative examples}
\label{tab:symbol_families} \\
\hline
Family code & \#symbols per family & Description & Example \\
\hline
\endfirsthead

\hline
Family code & \#symbols per family & Description & Example \\
\hline
\endhead

A & 11 & Circles & \includegraphics[width=1.2cm]{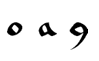} \\
B & 15 & Cross-overs & \includegraphics[width=1.2cm]{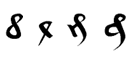} \\
C & 13 & Inverted S-shapes & \includegraphics[width=0.7cm]{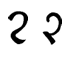} \\
D & 3 & 4-shape & \includegraphics[width=0.6cm]{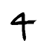} \\
E & 6 & Single \textit{i} & \includegraphics[width=0.6cm]{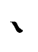} \\
F & 3 & Double \textit{i} & \includegraphics[width=1.5cm]{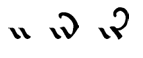} \\
G & 3 & Triple \textit{i} & \includegraphics[width=1.5cm]{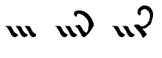} \\
H & 3 & Even more \textit{i}'s & \includegraphics[width=0.8cm]{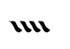} \\
J & 4 & Single \textit{c} & \includegraphics[width=0.6cm]{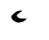} \\
K & 14 & Double \textit{c} (without plume) & \includegraphics[width=1.1cm]{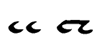} \\
L & 9 & Double \textit{c} (with plume) & \includegraphics[width=0.7cm]{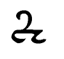} \\
M & 6 & Triple \textit{c} & \includegraphics[width=1.0cm]{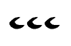} \\
N & 2 & Even more \textit{c}'s & \includegraphics[width=1.2cm]{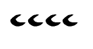} \\
P & 12 & Single vertical bar & \includegraphics[width=1.2cm]{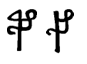} \\
Q & 11 & Double vertical bar & \includegraphics[width=1.2cm]{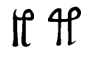} \\
R & 7 & Single vertical bar with pedestal (part 1 of 3) & \includegraphics[width=0.9cm]{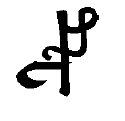} \\
S & 8 & Double vertical bar with pedestal (part 1 of 3) & \includegraphics[width=0.9cm]{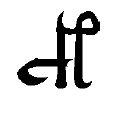} \\
T & 21 & Single vertical bar with pedestal (part 2 of 3) & \includegraphics[width=1.4cm]{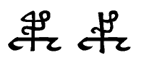} \\
U & 22 & Double vertical bar with pedestal (part 2 of 3) & \includegraphics[width=1.5cm]{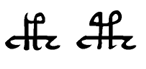} \\
V & 7 & Single vertical bar with pedestal (part 3 of 3) & \includegraphics[width=1.2cm]{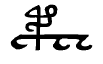} \\
W & 20 & Double vertical bar with pedestal (part 3 of 3) & \includegraphics[width=1.2cm]{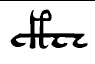} \\
X & 14 & Unclassifiable & --- \\
Z & 209 & Unknown & --- \\
\hline
\end{longtable}

\subsection{Word and Zipf distributions}

\begin{figure}[ht]
\centering
\includegraphics[width=0.75\textwidth]{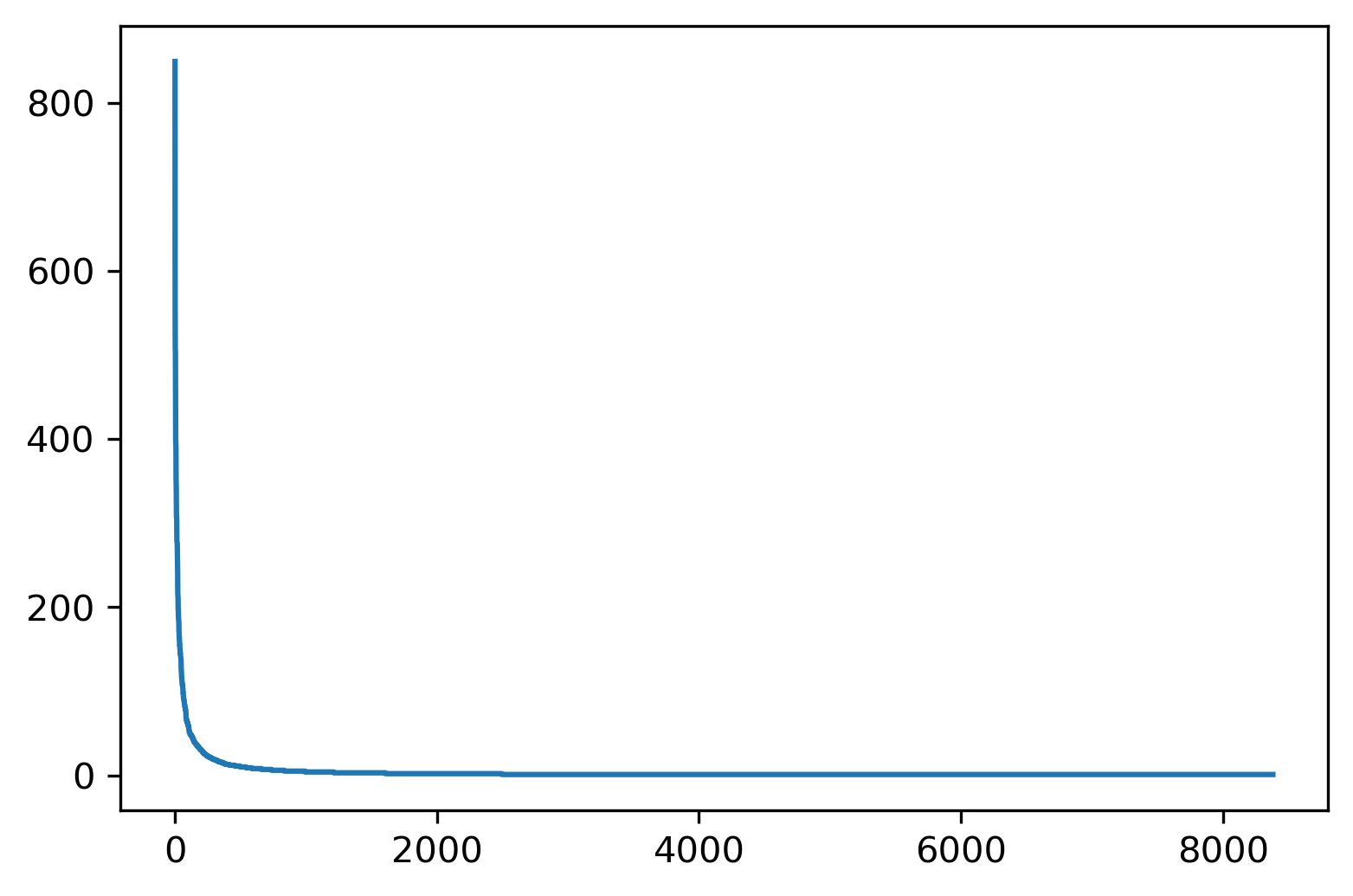}
\caption{Zipf distribution of word frequencies in the VM corpus.}
\label{fig:zipf_mvs}
  \Description{Log-log plot showing the Zipf distribution of word frequencies in the Voynich Manuscript corpus, with word rank on the x-axis and frequency on the y-axis.}
\end{figure}

We define a word as a contiguous sequence of glyphs that does not contain a delimiter. 
In Appendix (supplementary material p18) section "List of most frequent sequences of glyphs" presents a table, listing the most frequent words. 
The sequences B1A3G1 and A1B2 are the most frequent, occurring 849 and 558 times, respectively. 
The corpus contains a total of 8,383 distinct words, of which 5,886 are hapax legomena (words occurring only once), representing 70.2\% of the vocabulary. 
Figure~\ref{fig:zipf_mvs} shows the distribution of words ranked by decreasing frequency; the resulting curve closely follows Zipf’s power law (\cite{Zipf1935}), a characteristic feature of textual corpora in any natural language.

\subsection{Letter bigram and Word bigram distribution}

Theoretically, the 423 distinct symbols could generate 178,929 bigrams (423²) and 75,686,967 trigrams (423³). 
However, the subset of n-grams that are likely to carry meaningful combinatorial significance is considerably smaller.
We estimate that the number of potentially significant character bigrams is 2,209 (47²), while the number of potentially significant character trigrams is 103,823 (47³). 
From this subset, we identified 69 bigrams and 365 trigrams that occur across a large portion of the manuscript’s paragraphs.
In Appendix (supplementary material page18) section "List of most frequent bigrams and trigrams of glyphs" a table presents the frequencies of the 15 most common bigrams, whereas a table shows the frequencies of the 15 most common trigrams, both evaluated over a broad coverage of the manuscript’s paragraphs.

\subsection{Summary}

\begin{assumption}
Raw data contains 423 differents glyphs, 157,300 glyphs occurrences and 43,650 delimiter occurrence between glyph sequences. 
47 glyphs (having more or equal to 10 occurrences) correspond to 99.62\% of glyph occurrences. 
Number of contiguous glyphs (different simple words) is 8,383 distributed over 740 paragraphs over 116 folios (or 232 paper pages, or 204 pdf pages).
\end{assumption}

\begin{assumption}
The raw dataset contains 423 distinct glyphs, with a total of 157,300 glyph occurrences and 43,650 delimiters separating glyph sequences.  
A subset of 47 glyphs, each occurring at least 10 times, accounts for 99.62\% of all glyph occurrences.  
The number of contiguous glyph sequences (i.e., distinct simple words) is 8,383, distributed across 740 paragraphs spanning 116 folios (equivalent to 232 paper pages or 204 PDF pages).
\end{assumption}

\begin{assumption}
The number of potentially significant letter bigrams and trigrams is limited to 69 and 365, respectively.
\end{assumption}

\begin{assumption}
Word frequencies follow a Zipfian distribution.
\end{assumption}

\section{History of Ancient Herbal Medicine}
\label{sec:herbal}

See Appendix (supplementary material page1) section "History of Herbal Medicine" for further details.

\subsection{From 2500~BCE to 100~CE: Herbal Medicine}

Ancient medical knowledge developed through empirical observation intertwined with religious, magical, and symbolic belief systems. Early traditions—Mesopotamian, Egyptian, and Greco-Roman—established foundational concepts in medicine and pharmacology.\\
Mesopotamian medicine (ca. 2500 BCE) conceptualized disease as both a physical and a supernatural phenomenon. 
Physicians, known as asû(m), combined plant-based remedies, wound care, and minor surgical procedures with magical rituals, frequently invoking protective deities. 
Surviving medical tablets attest to advanced botanical knowledge and early attempts at plant classification.\\
Egyptian medicine, exemplified by the Ebers Papyrus (ca. 1550 BCE), likewise integrated therapeutic remedies with ritual practices. 
The papyrus records 877 prescriptions organized into thirty-three groups, predominantly based on plant substances, with treatments frequently accompanied by magical formulas. 
Medical knowledge was transmitted primarily through textual descriptions rather than visual representation.\\
The Greco-Roman tradition marked a shift toward systematic observation and classification. 
Theophrastus is regarded as the founder of botany, while Dioscorides’ De Materia Medica cataloged approximately 600 medicinal substances and exerted a lasting influence on medieval pharmacology. 
Illustrated manuscripts, most notably the Vienna Dioscorides, played a central role in the transmission of herbal knowledge throughout Europe and the Arab world.\\
Together, these traditions illustrate a gradual transition from ritualized medical practices toward more systematic and empirical approaches, while preserving a central focus on medicinal plants and a holistic conception of health.

\subsection{ Medieval Herbals – Latinus and Herbarium of Trento}

Between the Roman period and the fifteenth century, botanical knowledge was not compiled through collections of dried specimens, as in the modern herbarium, but primarily through illustrated representations. These compilations may therefore be described as artificial herbals, in which visual depiction rather than preserved plant material served as the principal medium for recording and transmitting botanical knowledge.\\
The Pseudo-Apuleius (fourth century CE) is a Latin herbal that describes 131 plants, providing their regional names, methods of collection, habitats, and medicinal uses (\cite{Pseudoapuleius2026}). 
The corpus largely reflects the flora of the Mediterranean world, encompassing southern Europe, the Balkans, Anatolia, and the Levant.\\
Latinus (Mainz, 1484) was a practical medical manual focused on the use of simple plants, in which plants were depicted with recognizable yet often schematic silhouettes. 
In medieval herbals more broadly, plants were sometimes rendered imaginatively to emphasize their medicinal or alchemical properties; roots and flowers could be highly stylized or symbolic, serving mnemonic purposes rather than strict morphological accuracy.\\
The earliest known surviving herbarium manuscript incorporating real, dried plant specimens dates to 1532–1533 and is attributed to Luca Ghini. 
By contrast, the late-fourteenth-century Herbarium of Trento, preserved at the Castello del Buonconsiglio, contains eighty-six illustrated medicinal plants and twenty-three pages of medical recipes, accompanied by annotations in Venetian dialect. 
Its watercolor illustrations combine real and imaginary plants, occasionally incorporating magical or cabalistic elements, and reflect a hybrid transmission of medical and botanical knowledge intended for both specialists and a broader vernacular audience.

\subsection{ Late Medieval and Renaissance Herbals}

During the Middle Ages (9th–15th centuries), between 200 and 300 medicinal plants are documented in herbals, monastic treatises, and pharmacopoeias (e.g., Hildegard of Bingen, Tacuinum Sanitatis, Circa instans), although only a portion were aromatic. 
The 15th-century northern Italian Erbario contains approximately 175 plant descriptions, presented in both Italian and Latin (\cite{ErbarioPenn2026}).\\
Manuscript Harley MS 585 (late 10th–early 11th century) contains translations of Pseudo-Apuleius and other medical texts, as well as the Leechbook (Lacnunga), incorporating marginal glosses, decorated initials, and illustrative drawings. 
Similarly, De herbarum virtutibus (Venetian-Verona, late 15th century) enumerates approximately 130 plants with medicinal properties, presenting illustrations that range from highly realistic (e.g., strawberry) to highly stylized (e.g., basil), reflecting both didactic and artistic conventions.\\
The Tractatus de Herbis (c. 1300, Southern Italy), also known as the Secreta Salernitana, associates plant names with illustrations to prevent errors in medical prescriptions (\cite{TractatusHerbis2026}). 
These collections contain between 500 and 900 entries of plant, mineral, and animal simples, disseminating the pharmacological tradition of the Salerno medical school throughout Europe. 
Early versions of the text are preserved in manuscripts such as Egerton MS 747.\\
Gale’s Herbarium (14th-century Germany) comprises approximately 800 plant illustrations alongside 40 depictions of doctors and patients, detailing the properties of each plant, the illnesses for which they were considered appropriate, and methods of preparation (\cite{GaleHerbarius1350}).

\section{Symbol representation}
\label{sec:symbolrep}

\subsection{Pure syllabic alphabet}

The Knot Alphabet was developed in 2010 by Sheldon Ebbeler, a linguist and analyst of verbal behavior. 
Technically, it is not a true alphabet; rather, it is an alphasyllabary, or abugida, in which each syllable is represented as a single unit. 
Within these syllabic units, the consonant component is primary, while the vowel component, though secondary, is still required. The system comprises approximately 70 distinct symbols to formalize syllabic usage of English.\\
As noted in the previous section, the VM corpus contains only 47 widely occurring symbols. 
Among these, however, only 17 are relatively rare and are characterized by one or two vertical bars.

\subsection{Cistercian alphabet}

The Cistercian numerical system, also referred to as Agrippa’s Notae Elegantissimae, is a medieval method of representing numbers, developed by European monks during the Middle Ages. 
It encodes numbers from 1 to 9,999 using a compact arrangement of strokes around a central vertical staff, rather than serving as a system for alphabetic characters \cite{cistercian2026}.\\
Surviving evidence of its use is scarce: only about two dozen manuscripts dating from the thirteenth to the fifteenth centuries are known, originating from regions ranging from England and Normandy to Italy and Sweden. 
Its material legacy also includes an astrolabe preserved at the Louvre Abu Dhabi, on which Cistercian numerals are engraved \cite{Astrolab1375}.
The system was not designed for arithmetic computation but rather for pragmatic purposes, including the notation of years, foliation, textual divisions, lists, indexes, concordances, Easter tables, and even musical notation. 
Its development can be understood in light of several practical and cultural factors: the need for rapid, space-efficient writing in monastic administration; the limited dissemination and lack of standardization of Arabic numerals in Western Europe at the time; and the intention to restrict access to sensitive information—such as inventories or records of donations—by employing a notation intelligible only to trained insiders.
These motivations must be situated within the historical context of the Cistercian Order, founded in 1098 at Cîteaux, which emphasized austerity, discipline, and administrative rigor. 
The order expanded rapidly across Europe during the twelfth and thirteenth centuries, eventually encompassing more than 650 abbeys and priories. 
The system ultimately fell into disuse by the sixteenth century, as Arabic numerals became standardized and dominant in administration, commerce, and science, offering greater simplicity for representing large numbers and enabling more efficient arithmetic operations.

\begin{table}[ht]
\centering
\begin{tabular}{c l l}
\hline
Cistercian symbol & Decimal Value      & Voynich Symbol \\
\hline
\includegraphics[width=0.6cm]{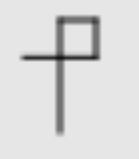} & 29        & \includegraphics[width=0.5cm]{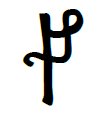} \\
\includegraphics[width=0.4cm]{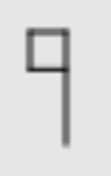}\includegraphics[width=0.38cm]{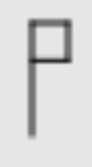} & 90, 9     & \includegraphics[width=0.5cm]{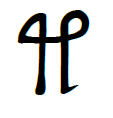} \\
\includegraphics[width=0.4cm]{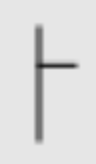}\includegraphics[width=0.38cm]{9.png} & 2, 9      & \includegraphics[width=0.5cm]{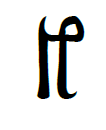}\\
\includegraphics[width=0.5cm]{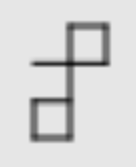} & 9029      & \includegraphics[width=0.5cm]{Ra.png} \\
\includegraphics[width=0.5cm]{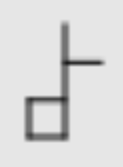}\includegraphics[width=0.4cm]{9.png} & 9002, 9   & \includegraphics[width=0.5cm]{S1.png} \\
\hline
\end{tabular}
\caption{Association between Cistercian Symbols and Voynich Symbols}
\label{tab:image_value_code}
\end{table}

\subsection{Summary}

The Voynich Manuscript has long attracted scholarly interest due to its use of entirely unfamiliar glyphs that do not correspond to any known writing system.

\begin{assumption}
In the Voynich Manuscript, some glyphs appear to have no clear origin in contemporary European scripts. The Cistercian numeral system may nevertheless have served as a source of inspiration for several Voynich glyph families, notably P, Q, R, S, T, U, V, and W, representing 8 out of the 21 identified glyph families.
\end{assumption}

\section{Phonetic decomposition: plant book comparison hypothesis}
\label{sec:phonetic}

\subsection{Manual Syllabic decomposition}

The law of diversification \cite{lqlDiversification} describes the tendency of languages to possess a small core of highly frequent elements (such as syllables or function words), while simultaneously allowing for unbounded expressive potential through the use of rare or infrequent units.\\

Below, we present a comparison between a paragraph from VM (Table~\ref{tab:voynich77r}), a Latin text on herbal medicine (Table~\ref{tab:plantago}), and a French text (Table~\ref{tab:camomille_bilingual}). 
For this comparison, the natural language texts are decomposed into their phonetic constituents (syllables) in order to relate them to the occurrences of symbols in the VM.

\begin{table}[htbp]
\centering
\small
\renewcommand\cellalign{tl}
\setlength{\tabcolsep}{4pt}

\begin{tabular}{|p{2cm}|p{8cm}|}
\hline
\textbf{Folio} & \textbf{Transliteration} \\
\hline

f77r.44 &
\makecell[tl]{
L1K2A2.D1J1P1K1J1B1A2.D1A1B2.K1K2B1A2. \\
D1A1Q1J1A3C1.K1K2A2.B2A1B2A2.B1A2
} \\
\hline

f77r.45 &
\makecell[tl]{
D1A1Q1A3G1.K1J1C2.B2K1J1A3C1.B1A3B2.K1C1. \\
C2B2,C2,A3F1.A1B2.C1A3G1.B2A1B1
} \\
\hline

f77r.46 &
\makecell[tl]{
B1A3G1.K1B2K1P1L1K2A2.Q2A3B2.K1J1A1B2. \\
B1A3B4.A3C2.A1Q2J1A2.B1A3G1
} \\
\hline

f77r.47 &
\makecell[tl]{
A2L1K2A2.B1A3B2.C2A3F2C2[A1:A3]B2. \\
C2A3B2.B1A3B2A1.Q2A3F1.B1A3C1A2.B1A3B2A1
} \\
\hline

\end{tabular}
\caption{Reference segment from the VM (folio 77r) used for analysis.}
\label{tab:voynich77r}
\end{table}

Table~\ref{tab:voynich77r} shows a total of 35 words comprising 123 symbols; the three most frequent symbols (A3, B2, and B1) together account for 37\% of all symbol occurrences.

\begin{table}[htbp]
\centering
\small
\begin{tabular}{p{0.3\textwidth} p{0.3\textwidth} p{0.3\textwidth}}
\hline
\textbf{Latin} & \textbf{French} & \textbf{English} \\
\hline
Plantago est herba humilis, foliis longis et angustis, viridibus, nervis tribus distinctis. Crescit in locis humidis et viarum marginibus. Utilis est ad vulnera sananda, si folia contrita superponantur. Succus eius cum melle mixtus tussim mitigat.
&
Le plantain est une herbe basse, aux feuilles longues et étroites, vertes, marquées de trois nervures. Elle pousse dans les lieux humides et sur les bords des chemins. Elle est utile pour guérir les blessures, si l’on applique ses feuilles écrasées. Son jus, mélangé avec du miel, apaise la toux.
&
The plantain is a low-growing herb with long, narrow, green leaves marked by three distinct veins. It grows in damp places and along the edges of paths. It is useful for healing wounds when its crushed leaves are applied. Its juice, mixed with honey, soothes coughs.
\\
\hline
\end{tabular}
\caption{Description of \emph{Plantago}: Latin source text with French and English translations}
\label{tab:plantago}
\end{table}

From Table~\ref{tab:plantago}, a reference Latin text was selected for its relevance to herbal medicine. This text was decomposed into elementary syllabic units. 
We first constructed a dictionary comprising 34 words, 214 letters, and 81 syllables. 
Among the 81 syllables identified (65 of which are unique), 17 occur frequently, accounting for 25\% of the total syllable occurrences, and two syllables appear with a frequency of three.
The three most frequent syllables (\emph{bus}, \emph{mis}, and \emph{a}) together account for 9.8\% of all syllable occurrences, whereas the three most frequent letters (\emph{i}, \emph{s}, and \emph{t}) collectively represent 35.5\% of the total letter frequency.

\begin{table}[htbp]
\centering
\small
\begin{tabular}{p{0.45\textwidth} p{0.45\textwidth}}
\hline
\textbf{Français} & \textbf{English} \\
\hline
La camomille est une plante médicinale utilisée depuis l’Antiquité. Ses fleurs, en infusion ou décoction, calment les troubles digestifs, favorisent le sommeil, apaisent l’anxiété et possèdent des propriétés anti-inflammatoires et relaxantes naturelles. 
&
Chamomile is a medicinal plant used since Antiquity. Its flowers, in infusion or decoction, soothe digestive disorders, promote sleep, calm anxiety, and have natural anti-inflammatory and relaxing properties. \\
\hline
\end{tabular}
\caption{ Text about Medicinal properties of chamomile}
\label{tab:camomille_bilingual}
\end{table}

The French text in Table~\ref{tab:camomille_bilingual} contains 41 words, 241 letters, and 82 syllables; the three most frequent syllables—\emph{le}, \emph{an}, and \emph{é}—together account for 13\% of all syllable occurrences.\\
In Mandarin Chinese, there are 21 consonants and 36 vowels or finals, yielding approximately 57 basic phonemic units. 
These units combine to form around 400 distinct syllables, each of which may carry one of four tonal variations, resulting in roughly 1,200 possible syllabic forms.\\
By comparison, \cite{zornigAltmann1993} report 610 syllable types in Indonesian, while \cite{bony1922} estimate that French has approximately 1,430 distinct syllables. 
According to the Académie du français langue étrangère, 348 fundamental syllables are regarded as distinct.\\

\subsection{Known-Unknown symbol Matrix model}

\cite{Lasry04032023} deciphered a set of previously unknown ciphered letters written by Mary Stuart in 1583, after more than 440 years. 
The cipher employed in this correspondence used a set of approximately 191 distinct graphical symbols. These symbols formed part of a homophonic cipher system, in which multiple symbols could represent the same plaintext letter, and in some cases common words or proper names.
Decipherment can be framed as the problem of determining a mapping between known plaintext elements (rows) and unknown ciphertext symbols (columns), represented as a matrix. 
The objective of the deciphering process is to resolve the entries of this matrix in order to recover the correct symbol-to-letter correspondences.
Some encoded symbols correspond to well-known names from the period, including named entities such as places and personal names.  
Using a similar approach, \cite{desset2022decipherment} deciphered a script approximately 5,000 years old—Proto-Elamite—from Iran by solving the correspondence between phonemes and previously unknown symbols.

\subsection{Reference texts}

We aim to compare the phonetic properties of a reference text drawn from an ancient manuscript on medicinal plants with those of a Voynich Manuscript passage of comparable length.
We consider an example of a plant description, namely \emph{Solago Minor}, as documented in the \emph{Herbarius} (\cite{Pseudoapuleius10};  see Appendix (supplementary material page21) section "Phonetic decomposition". 
Table~\ref{tab:solago} presents the corresponding translations in French and English.
As a reference, we randomly selected a paragraph from the Voynich Manuscript of comparable size for comparison of phonetic properties ; see Appendix (supplementary material page21) section "Phonetic decomposition". 
The passage comprises 11 lines and contains 28 distinct symbols.\\

The Latin reference text consists of 15 lines and 88 words. The four longest words are \emph{scorpionidicum} (14 letters), \emph{lubricitatem} (12 letters), \emph{contusionem} (11 letters), and \emph{Solagominor} (11 letters). 
The Voynich reference passage comprises 11 lines and 71 tokens. Within this passage, we observe three longest contiguous symbol sequences: A1Q1K2A1B1A3G1 (7 symbols), D1J1Q1J1A1B1A2 (7 symbols), and K1J1A1B1A2 (5 symbols).

\begin{table}[ht]
\centering
\renewcommand{\arraystretch}{1.3}
\begin{tabular}{|p{6cm}|p{8cm}|}
\hline
\textbf{Français} & \textbf{English} \\
\hline
Elle pousse partout. & It grows everywhere. \\
\hline
Pour soigner les glaires, l’herbe Solago mineure, séchée et réduite en poudre, donnée à boire avec de l’eau chaude, fait tomber les mucosités et les expulse. &
To treat phlegm, the herb *Solago minor*, dried and reduced to powder, is given to drink with hot water; it causes the mucosities to loosen and be expelled. \\
\hline
On la cueille en été. & It is harvested in summer. \\
\hline
Des vertus de l’herbe Solago mineure. & On the virtues of the herb *Solago minor*. \\
\hline
Elle est utile contre une affection appelée libricinem (probablement un trouble gynécologique ou intestinal). &
It is useful against an ailment called *libricinem* (probably a gynecological or intestinal disorder). \\
\hline
Herbe Solago majeure. & Herb *Solago major*. \\
\hline
De même, la Solago majeure, par ses deux formes et ses propriétés médicinales. &
Likewise, *Solago major*, with its two forms and its medicinal properties. \\
\hline
Elle soigne aussi les piqûres de scorpions. &
It also treats scorpion stings. \\
\hline
Pour les morsures de serpent et les piqûres de scorpion, l’herbe Solago majeure, séchée et réduite en une poudre très fine, est donnée à boire dans du vin. &
For snake bites and scorpion stings, the herb *Solago major*, dried and reduced to a very fine powder, is given to drink in wine. \\
\hline
On applique aussi directement la plante sur la contusion. &
The plant is also applied directly to the bruise. \\
\hline
\end{tabular}
\caption{French–English translations of descriptions of \emph{Solago minor} and \emph{Solago major} (archived at BNF Gallica, Paris \cite{Pseudoapuleius10}, folios 63r and 63v, see  picture - Appendix (supplementary material page 21) section "Original document".}
\label{tab:solago}
\end{table}

In the next step, we use a Latin text as reference as described in Table~\ref{tab:solago}(see  pictures in Appendix (supplementary material page 21) section "Original document" from \cite{Pseudoapuleius10}). we decompose each word into a sequence of elementary phonetic symbols using a composition matrix (see Table In Appendix (supplementary material page21) section "Vowel–Consonant Matrix" . 
In total, the decomposition yields 101 distinct elementary phonetic units. For example, the symbol \emph{s28} corresponds to the sound combination ``s+e''. 
These elementary units may represent either a vowel or a single consonant.

\begin{align}
\text{To convert the French word ``maison'' we obtain:} \nonumber\\
m + e &\rightarrow s68 \\
z + o &\rightarrow s100 \\
n\;(\text{standalone consonant}) &\rightarrow s16 \\
\textbf{maison} &\rightarrow \textbf{s68 · s100 · s16} \nonumber\\[1ex]
\text{To convert the English word ``house'' we obtain:} \nonumber\\
h + a &\rightarrow s32 \\
h + u &\rightarrow s36 \\
s\;(\text{standalone consonant}) &\rightarrow s7 \\
\textbf{house} &\rightarrow \textbf{s32 · s36 · s7} \nonumber
\end{align}

We aim to estimate the number of phonetic units that a text can produce across different languages and to compare the number of units required in each case. 
To this end, we conducted decomposition experiments across ten European languages commonly used during the Middle Ages—namely Latin, French, English, Flemish, German, Spanish, Portuguese, Italian, Catalan, and Provençal. 
We additionally included Arabic and Hebrew, which were also widely used during this period.
We also include Mandarin Chinese in the comparison, as it represents a particularly compact language. 
For Arabic, Hebrew, and Chinese, the phonetic decomposition matrix described above is not directly applicable. 
Consequently, for Arabic and Hebrew we estimate phonetic complexity by counting consonantal units, while for Chinese we base the count on the units of Pinyin transcription.

\begin{figure}[ht!]
\centering
\begin{tikzpicture}
\begin{axis}[
    xbar,
    width=12cm,
    height=10cm,
    xmin=0,
    ylabel={Languages},
    xlabel={Number of distinct symbols / syllables},
    symbolic y coords={
        Medieval Hebrew,
        Medieval Arabic, Voynich MS, Medieval Latin, Medieval French, Medieval English,
        Medieval Spanish, Medieval Portuguese, Medieval Italian,
        Medieval Flemish, Medieval German, Medieval Catalan, Medieval Provencal,
        Medieval Chinese
    },
    ytick=data,
    nodes near coords,
    nodes near coords align={horizontal},
    bar width=8pt,
    enlarge y limits=0.05,
]
\addplot coordinates {
    (22,Medieval Hebrew)
    (26,Medieval Arabic)
    (28,Voynich MS)
    (60,Medieval Latin)
    (42,Medieval French)
    (74,Medieval English)
    (63,Medieval Spanish)
    (65,Medieval Portuguese)
    (62,Medieval Italian)
    (42,Medieval Flemish)
    (47,Medieval German)
    (56,Medieval Catalan)
    (41,Medieval Provencal)
    (53,Medieval Chinese)
};
\end{axis}
\end{tikzpicture}
\caption{Distinct phentic symbol inventory across medieval languages}
\Description{Horizontal bar chart comparing the number of distinct phonetic symbols or syllables used in several medieval languages and the Voynich Manuscript. The Voynich Manuscript shows 28 symbols, while other languages range approximately from 22 to 74 symbols.}

\end{figure}
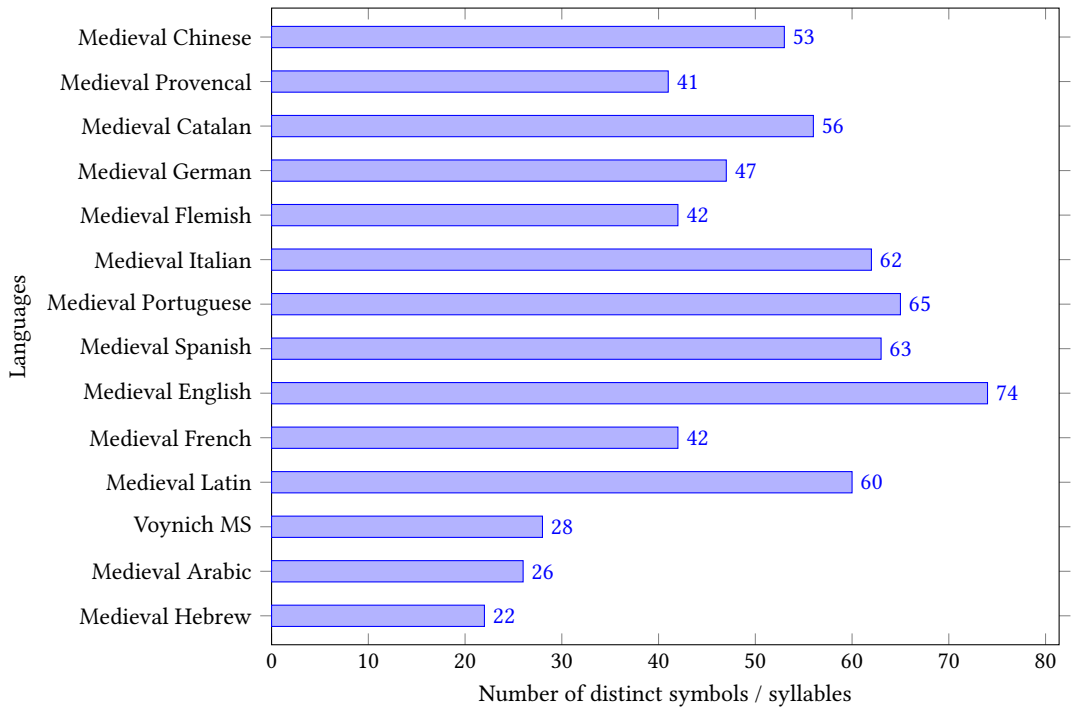

The longest words in Hebrew is shown on figure-\ref{fig:hebreww} below:

\begin{figure}[H]
    \centering
    \includegraphics[width=0.2\textwidth]{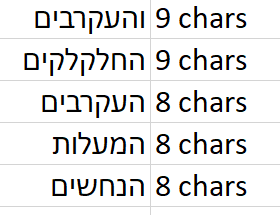} 
    \caption{ longuest words in Hebrew}
    \label{fig:hebreww}
        \Description{Bar chart showing the longest word lengths observed in a Hebrew text corpus, illustrating the distribution of word sizes in the dataset.}

\end{figure}

The longest word in Arabic is:

(al-mustaḥḍarāt) → 11 letters. (the remedies)   

The longest words in Mandarin Chinese are: \begin{CJK}{UTF8}{gbsn}大蘇拉戈草亦主蛇咬 \end{CJK}Dà Sūlāgē cǎo yì zhǔ shé yǎo 9, \begin{CJK}{UTF8}{gbsn}以小蘇拉戈草乾之\end{CJK} yǐ xiǎo Sūlāgē cǎo gān zhī 8, \begin{CJK}{UTF8}{gbsn}論小蘇拉戈草之德 \end{CJK}Lùn xiǎo Sūlāgē cǎo zhī dé 8

\subsection{Summary}

The Voynich Manuscript script cannot be purely syllabic. The number of commonly used distinct symbols is approximately 47, whereas even compact languages such as Chinese—and certainly European languages—require several hundred distinct syllables. Moreover, in short passages, the most frequent Voynich symbols behave more like letters than syllables, based on their distributional properties.

\begin{assumption}
In the VM, the distribution of symbol sequences exhibits letter-like behavior, whereas the distribution of word lengths is more consistent with that of syllables.
\end{assumption}

\begin{assumption}
In the VM, tokens appear to behave similarly to those of consonantal writing systems such as Hebrew or Arabic, rather than to those of East Asian languages such as Mandarin or Indo-European languages.
\end{assumption}

\section{Language encoding: most frequent letter hypothesis}
\label{sec:frequentletter}

\subsection{Middle-age encoding systems}

\begin{figure}[ht]
    \centering
    \includegraphics[width=0.5\textwidth]{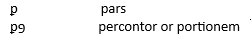} 
    \caption{ \emph{pars} and \emph{percontor} symbols}
    \label{fig:percontor}
        \Description{Image showing two glyphs labeled as the \emph{pars} and \emph{percontor} symbols, used for comparison of their shapes and structure.}

\end{figure}

In a medical manuscript from Monte Cassino (11th–12th centuries), certain plant names are encoded using an alphabetical shift, representing a proto–Caesar cipher. For example, the letter “B” is substituted for “A,” transforming “Aloe” into “Blpf”. This technique was employed to obscure the identity of rare or valuable plants. \\
Similarly, in the case of the \emph{Herbarius pictus} (Germany, 15th century), certain plants—such as \emph{Cannabis}, \emph{Papaver}, and \emph{Hyoscyamus}—are recorded using deliberately obscure abbreviations. This practice likely reflects an intention to prevent the disclosure of psychoactive substances to unauthorized readers. \\
In illustrated herbals associated with the School of Salerno during the 13th century, remedies aimed at combating demons or addressing so-called “female diseases” frequently employed words that were transformed into acronyms or written in non-Latin characters. Example: \emph{Ruta} is abbreviated as “R\textasciitilde a” with a cryptic bar, or replaced by a cross followed by an initial.\\
In certain manuscripts of the \emph{Herbarius Latinus} (15th century), dosages and recipe instructions are abbreviated through the use of invented symbols (see, for example, Figure~\ref{fig:percontor}).
This practice served to conceal the exact composition of the recipe, as the monks deliberately provided only partial or incomplete clues.

The \emph{Tabula recta} introduced by Johannes Trithemius in the late 15th century reflects the work of the Benedictine abbot (1462–1516), whose \emph{Steganographia} (1499) combines cryptographic techniques with elements of angelic magic \cite{Tritheme2026}.
Trithemius’ \emph{Tabula recta} is a square table of alphabets in which the first row consists of the Latin alphabet and each subsequent row is obtained by shifting the preceding row one position to the left. Although developed slightly later, this system builds upon a medieval monastic tradition of secret writing.
Despite long-standing awareness of the weaknesses inherent in simple substitution ciphers, and notwithstanding the earlier experiments of Alberti, Porta, and Trithemius, no genuinely new cryptographic method emerged between the time of Caesar and the 16th century. Such a method would have needed to be both practically secure by contemporary standards and straightforward to use.
Blaise de Vigenère, born in 1523, developed a novel method for encrypting messages that remained resistant to cryptanalysis for three centuries.  
Vigenère was a man of many talents: he was an alchemist, writer, and historian, and also served as a diplomat for the Dukes of Nevers and the kings of France.
In 1586, he published his \emph{Traicté des chiffres ou Secrètes manières d'écrire}.  
Vigenère’s innovation consisted in applying a Caesar cipher with a shift that varies from letter to letter \cite{Vigenere2026}.
To implement this method, a table consisting of 26 alphabets is used, each written in order, with every successive row shifted one position to the left relative to the previous row.  
To encode a message, a key—a word of arbitrary length—is selected.
The key is then written beneath the message to be encoded, repeated as many times as necessary so that each letter of the message has a corresponding letter of the key.  
To perform the encoding, one locates the intersection in the table between the row corresponding to the message letter and the column corresponding to the key letter.
For example, suppose we wish to encode the text ``CRYPTOGRAPHIE'' using the key ``MATHWEB''.  
First, the key is written beneath the text, repeating it as necessary.  
To encode the letter ``C'', the corresponding key letter is ``M''. 
We then locate the intersection in the matrix of the row beginning with ``C'' and the column beginning with ``M''.
This matrix is made by the rule: Cipher=(Message+Key)mod26\\
Example C + M → 2 + 12 = 14 → O\\
Appendix (supplementary material page 18) section "Conversion with Most Frequent Letters" give details of the transcription. Final Ciphertext  is "ORRWPSHDAIOEI". \\

The encryption methods discussed in this section are noteworthy; however, they do not involve the transformation of one alphabet into an entirely unknown alphabet.

\subsection{Text Conversion via Frequent-Letter Mapping}

Let us assume that a given Voynich text represents a conversion of a text from a known language.  
In that case, the most frequent letters of the original language would be mapped to symbols in the new script, such that the most frequent letters correspond to the most frequently occurring symbols in the Voynich text.
If this assumption holds, the reverse conversion should reveal the original language.  
We examine thirteen commonly used languages in medieval Western Europe and identify the twelve most frequent letters in each.  
These letters are then mapped to the twelve most frequent symbols in the VM (see table-\ref{tab:12freqglyph}), and the resulting mapping is used to convert the text in an attempt to identify recognizable words.

\begin{table}[ht]
\centering
\renewcommand{\arraystretch}{1.2} 
\begin{tabular}{c c c}
\toprule
Rank & Voynich Symbol & Number of Occurrences \\ 
\midrule
1  & A1 & 25,596 \\
2  & A2 & 17,872 \\
3  & A3 & 14,828 \\
4  & B1 & 13,217 \\
5  & K1 & 11,044 \\
6  & B2 & 10,659 \\
7  & J1 & 10,218 \\
8  & Q1 & 10,104 \\
9  & C1 & 6,819 \\
10 & Q2 & 6,017 \\
11 & D1 & 5,425 \\
12 & L1 & 4,533 \\
\bottomrule
\end{tabular}
\caption{The 12 most frequent symbols in the Voynich manuscript}
\label{tab:12freqglyph}
\end{table}

Below is a passage containing the most frequent letters of VM from folio 87 (see Table~\ref{tab:vmpassage}).  
We use large language model (LLM) prompting in the following manner: first, we request the most frequent letters in thirteen widely used medieval languages, namely French, English, Spanish, Portuguese, Italian, Catalan, Dutch, German, Provençal, Hebrew, Latin, Greek, and Arabic.
Next, a second prompt instructs the model to convert each of the frequent letters in the passage according to their respective ranks.  
Finally, we ask the model to identify any recognizable words resulting from this conversion.  
This procedure, consisting of three queries per language, yields a total of 39 queries submitted to ChatGPT version 5.0. \cite{chatgpt2026}.

\begin{table}[!ht]
\centering
\renewcommand{\arraystretch}{1.2} 

\begin{adjustbox}{max width=\textwidth}
\begin{tabular}{|>{\RaggedRight\arraybackslash\sloppy\textit}p{0.95\textwidth}|}
\hline
\seqsplit{ A2Q1J1.A1Q1K2C2.A1B2K1[J1:E1]B1[A3:A2].A1Q1A3B2,A2.A1B2.A1Q2A1B1A3C1.L1J1Q1A3B2<->A1B2Q1A3C1.K1J1A2Q1A3C1.L1A3B2.D1A1Q1A3B2.A1Q1[K1:A3]C1.A1Q2A3C1.A1Q2A1B1A2.A1Q2A1B1A2.A1Q2A1B1A2.A1Q2A3C2.K1P2A1B2.A1Q1L1A2.A1Q1[A2:Z1].A1Q1A3C1.A3C1.A1Q1J1B1A2.A1Q1J1B1A2.A1Q1K1J1B1A2.K1J1Q1K1J1B1A2.A1P2K1B1A1C1.D1A1Q2K1B1A2.Q1A1C1.K1B1A3C1,K1J1Q1A1B2Q1A1T3}
\\
\hline

\end{tabular}
\end{adjustbox}
\caption{Voynich Manuscript passage (VM folio 87)}
\label{tab:vmpassage}
\end{table}

\begin{figure}[ht]
    \centering
    \includegraphics[width=1.1\textwidth]{ 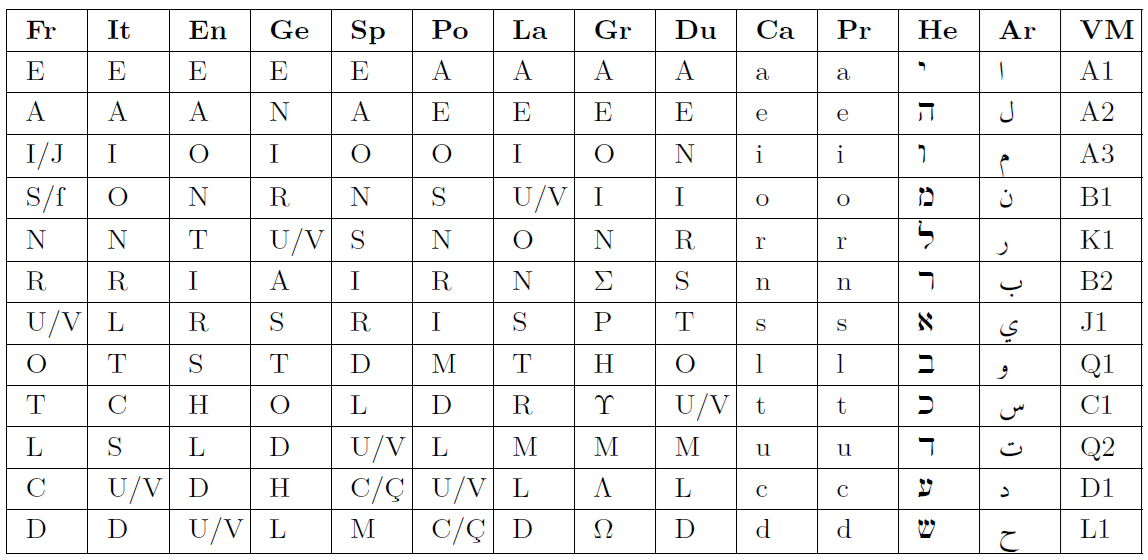} 
\caption{The twelve most frequent letters from medieval scripts and their corresponding Voynich symbols. Abbreviations: \textbf{Fr} – medieval French, \textbf{It} – medieval Italian, \textbf{En} – medieval English, \textbf{Ge} – medieval German, \textbf{Sp} – medieval Spanish, \textbf{Po} – medieval Portuguese, \textbf{La} – medieval Latin, \textbf{Gr} – medieval Greek, \textbf{Du} – medieval Dutch, \textbf{Ca} – medieval Catalan, \textbf{Pr} – medieval Provençal, \textbf{He} – medieval Hebrew, \textbf{Ar} – medieval Arabic, and \textbf{VM} – Voynich script.}    \label{fig:feqlet}
    \Description{Table-style figure comparing the twelve most frequent letters in several medieval languages with their visually corresponding symbols in the Voynich Manuscript. Each column represents a language and each row shows one of the most frequent characters.}

\end{figure}

\subsection{Summary}

\begin{assumption}
Converting the VM into a common Western European language will produce sequences of words that are not comprehensible.
\end{assumption}

\section{Image deciphering: real medicinal plant hypothesis}
\label{sec:imagemed}

\subsection{ Plant mapping }

Plants play a central role in the Voynich Manuscript. Of the 232 pages it contains, 129 depict plants accompanied by what appear to be descriptive elements, representing at least 63\% of the manuscript’s total content. \\

We extracted a total of 129 plant images from the Voynich Manuscript. A preliminary observation is that, among extant medieval herbals, the number of plant illustrations is rarely fixed and typically varies widely, ranging approximately from 100 to 500 entries. Notably, one of the most well-known herbals, the Pseudo-Apuleius, contains 131 plant descriptions, each accompanied by an illustration. This remarkably specific number is strikingly close to that observed in the Voynich Manuscript, suggesting a potentially meaningful correspondence rather than a coincidental similarity.\\

Scientific plant names are widely disseminated through biodiversity platforms such as GBIF \cite{Yesson2007GBIF}. However, the standardization and validation of these names largely occurred only in the past century. During the Middle Ages, plants were commonly referred to by vernacular or pre-Linnaean names that have evolved substantially over time. Consequently, establishing a correspondence between medieval plant names and modern scientific nomenclature is far from straightforward. In her doctoral thesis, Pradel \cite{Pradel2013} undertook a systematic mapping of plant names found in the Pseudo-Apuleius herbal to their contemporary scientific equivalents, providing a valuable starting point for such comparative work. Nevertheless, it remains unproven that the plants depicted in the Voynich Manuscript correspond entirely, or even predominantly, to those described in the Pseudo-Apuleius tradition. \\

\begin{figure}[htbp]
  \centering
  \begin{subfigure}{0.48\linewidth}
    \centering
    \includegraphics[width=\linewidth]{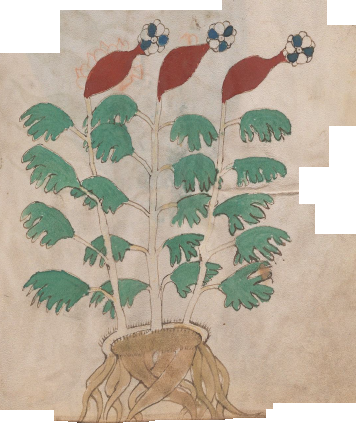}
    \caption{ page 106 (folio 55r)}
    \label{fig:p106_55r}
  \end{subfigure}
  \hfill
  \begin{subfigure}{0.48\linewidth}
    \centering
    \includegraphics[width=\linewidth]{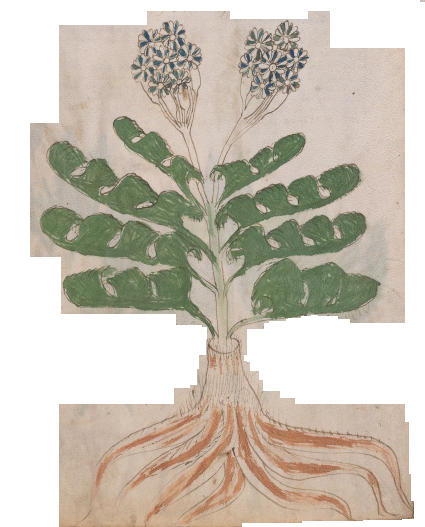}
    \caption{  page 75 (folio 39v)}
    \label{fig:p75_39v}
  \end{subfigure}
  \caption{Voynich plant pictures with symmetry.}
  \label{fig:p75_39v-p75_39v}
    \Description{Two images of plant illustrations from the Voynich Manuscript showing symmetrical designs. The first is from page 106 (folio 55r), the second from page 75 (folio 39v).}

\end{figure}

The plant representations in the Voynich Manuscript are characteristic of medieval botanical traditions, comparable to those found in contemporary herbals of the same period. These illustrations are typically highly stylized: while major plant organs are depicted, their proportions are often inaccurate, and leaf venation is rarely rendered with anatomical precision. In certain cases, symbolic elements—such as serpents or scorpions—are incorporated into the imagery, possibly to signify the toxic or dangerous properties of specific plant parts. \\
We observed that, among the plant illustrations in the Voynich Manuscript, approximately 58.6\% depict flowers as blue—a hue that is extremely rare in nature. This suggests that the author employed blue as a stylistic device, a creative liberty that was likely acceptable in the artistic conventions of the period. Similarly, the venation of leaves is often omitted; this absence does not imply that leaf structure was considered unimportant, but rather reflects the illustrative norms of the time. Moreover, some plants display symmetries that are virtually impossible to encounter in nature (Figure~\ref{fig:p75_39v-p75_39v}a), or exhibit phyllotactic patterns across multiple stems that are highly unusual and occur naturally only under specific developmental stages (juvenile versus adult, opposite versus alternate leaf arrangement) (Figure~\ref{fig:p75_39v-p75_39v}b). Therefore, any comparison between manuscript illustrations and real-world botanical forms must take into account the stylistic liberties of the author and the illustrative conventions of the era.

\subsection{ Mapping Methodology }

We used ChaGPT (LLM - Large Language modem)(\cite{Reuters2025}), Gallica WISE (SLM: Small Language Model)(\cite{gallica2026}) and Google image retrieval (HBIR: Hybrid-based Image Retrieval Search Engine)(\cite{google_homepage}).\\
Google image retrieval Ranking relies on text-based retrieval (Web page authority, relevance, and freshness, Textual context: captions, alt text, surrounding content.,User intent and query relevance) and content-based retrieval algorithm (Visual features extracted via CNNs / Vision Transformers). BNF Gallica WISE MERU Project  developping WISE is an AI-powered search engine for images and videos. WISE use multimodal OpenCLIP model. It indexes 1,225,136 images from Gallica. OpenCLIP uses Transformer architectures as part of its core design.\\

We assume we need to find an hypothesis each picture of the Voynich (N=129).

\begin{enumerate}
    \item \textbf{Reference Mapping:}  
    Map each Voynich plant illustration to its corresponding historical reference in Pseudo-Apuleius.  The total mappings is 131 plants. Output from this stage is a list we call HList (herbarius list of plant)( see Appendix (supplementary material page 6) section "Mapping from Pseudoapuleius and Scientific names").
    
    \item \textbf{LLM-Based Name Guessing with ChatGPT :}  
    For each 1:N picture, we prompt ChatGPT to make suggestion according organ representations or morphological trait (flower, stem, root,leaf). And we also ask for anomalies.
    We generate  2*N queries  ( see Appendix (supplementary material page 11) section "ChatGPT prompting for Plant identification" ).
    
    \item \textbf{Image-Based Verification:}  
    For each 3 (in average) candidate names from previous stage look on google image retrieval searching by name.
    We generate  3*N queries . Return a match.\\

    \item \textbf{Gallica WISE Search:}  
    For each 1:N picture, query with search engine. Look at the 150 images retrieved in the ranking. \\
    We generate  N queries. Return a match.\\

    \item \textbf{for last 50 unsolved pictures,LLM-Based Name Guessing with ChatGPT and  HList:}  
    For each 1:50 picture, we prompt ChatGPT to make suggestion according organ representations or morphological trait (flower, stem, root,leaf). And we also ask for anomalies. \\
    We generate  50 queries ( see Appendix (supplementary material page 11) section "ChatGPT prompting for Plant identification" ). 
    Make a checking on Google Image Retrieval.\\
    Return a match.\\

\end{enumerate}

\begin{table}[ht!]
\centering
\begin{tabular}{|l|c|}
\hline
\textbf{Method} & \textbf{Number of Queries} \\
\hline
Name Guessing (ChatGPT) & $2 \times N$ \\
Image-Based Verification & $3 \times N$ \\
Image Search (Gallica WISE) & $N$ \\
Last Pictures (ChatGPT, HList) & 50 \\
Image-Based Verification & 50 \\
total & 874 ($2 \times N$+100) \\
\hline
\end{tabular}
\caption{Summary of queries.}
\label{tab:queries}
\end{table}

As summarized on the Table~\ref{tab:queries} globally 874 queries have been used But we need to consider also Each image-based verification involved examining several dozen images retrieved by the search engine (\~10*((3*N)+50)+150*N concretely \~23,720) image consultations).

\subsection{ Equivalence between pseudoapulius plant name and actual }

Our objective is to annotate each illustration (see an example in Appendix (supplementary material page 12) section "Mapping Table (plant,scientific name)" with a corresponding scientific name, following the methodology described in the previous section. Of the 129 images examined, we successfully resolved the annotation task for 93 \% of the corpus. The following images could not be confidently annotated: picture 5 (folio 3v), picture 26 (folio 15r), picture 39 (folio 21v), picture 45 (folio 24v), picture 52 (folio 28r), picture 53 (folio 28v), picture 75 (folio 39v), picture 77 (folio 40v), and picture 106 (folio 55r).

\begin{figure}[htbp]
  \centering
  \begin{subfigure}[t]{0.32\textwidth}
    \centering
    \includegraphics[width=\textwidth]{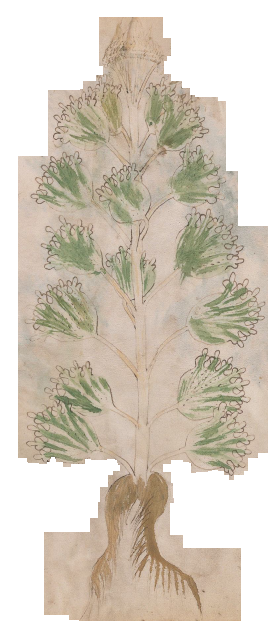}
    \caption{Page 78 illustration in the Voynich Manuscript \cite{VoynichOriginal2026}}
    \label{fig:voynich_p78}
  \end{subfigure}\hfill
  \begin{subfigure}[t]{0.32\textwidth}
    \centering
    \includegraphics[width=\textwidth]{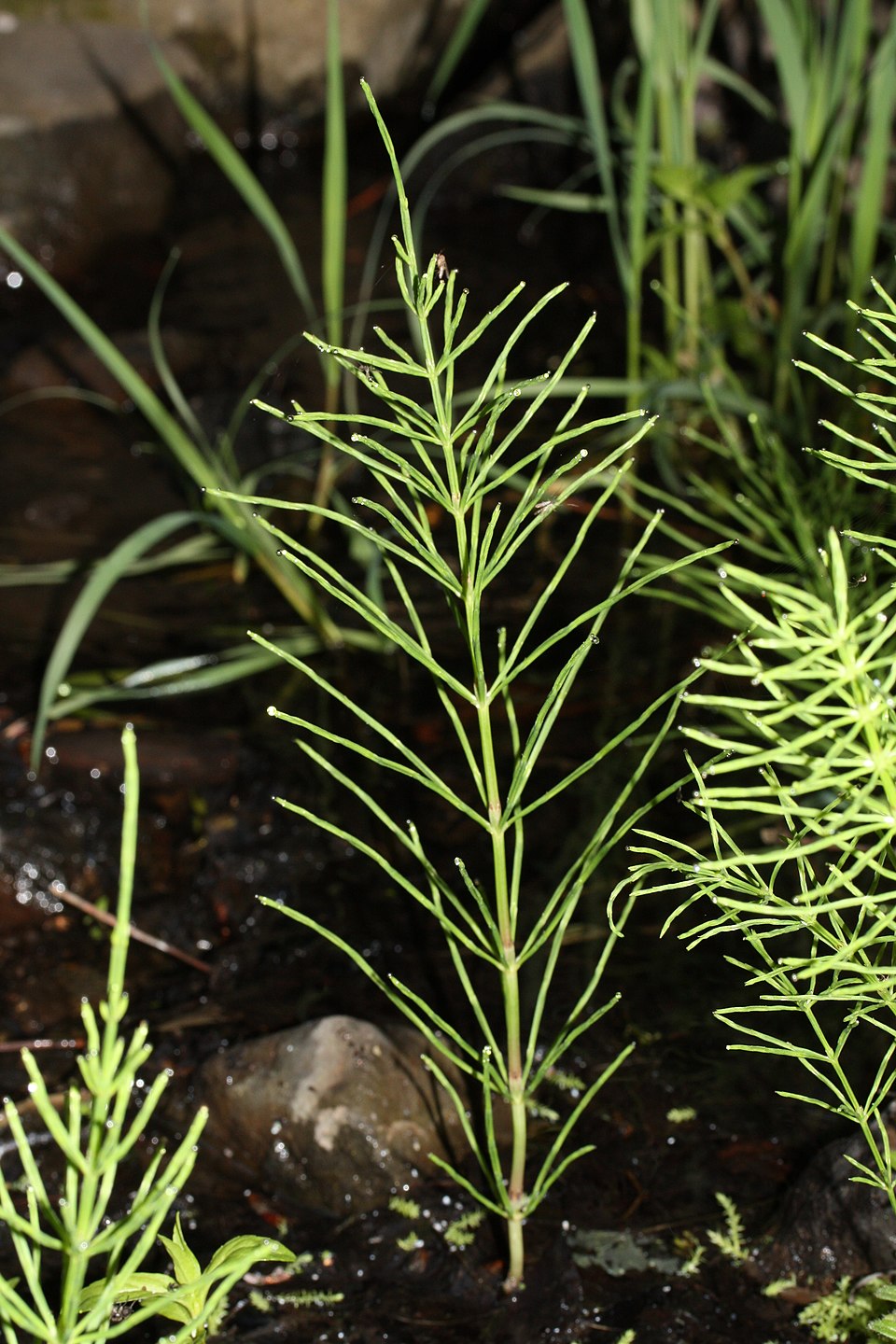}
    \caption{\textit{Equisetum arvense} (source: Wikipedia)}
    \label{fig:equisetum}
  \end{subfigure}\hfill
  \begin{subfigure}[t]{0.32\textwidth}
    \centering
    \includegraphics[width=\textwidth]{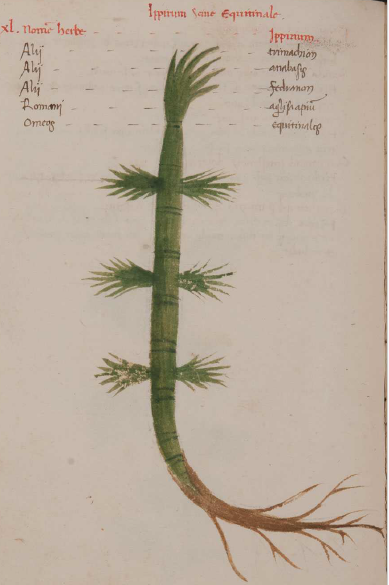}
    \caption{Illustration from \textit{De herbarum virtutibus}, 15th-century Pseudo-Apuleius manuscript \cite{DeHerbarumVirtutibus_Ar26}}
    \label{fig:pseudo_apuleius}
  \end{subfigure}
  \caption{Comparison of plant illustrations from the Voynich Manuscript, a modern botanical reference, and a medieval Pseudo-Apuleius herbal.}
  \label{fig:three-images}
    \Description{Three side-by-side images: (a) a plant illustration from page 78 of the Voynich Manuscript, (b) a modern photograph of \textit{Equisetum arvense}, and (c) a 15th-century illustration from the Pseudo-Apuleius herbal \textit{De herbarum virtutibus}, showing differences in artistic style and plant depiction across sources.}

\end{figure}

Table \ref{fig:three-images} presents an example of the annotation process. The complete set of annotations, including all image indices, is provided in Appendix (supplementary material page 6) section "Plant picture".

\subsection{ Area coverage of plant hypotheses }

\begin{figure}[htbp]
  \centering
  \includegraphics[width=0.5\linewidth]{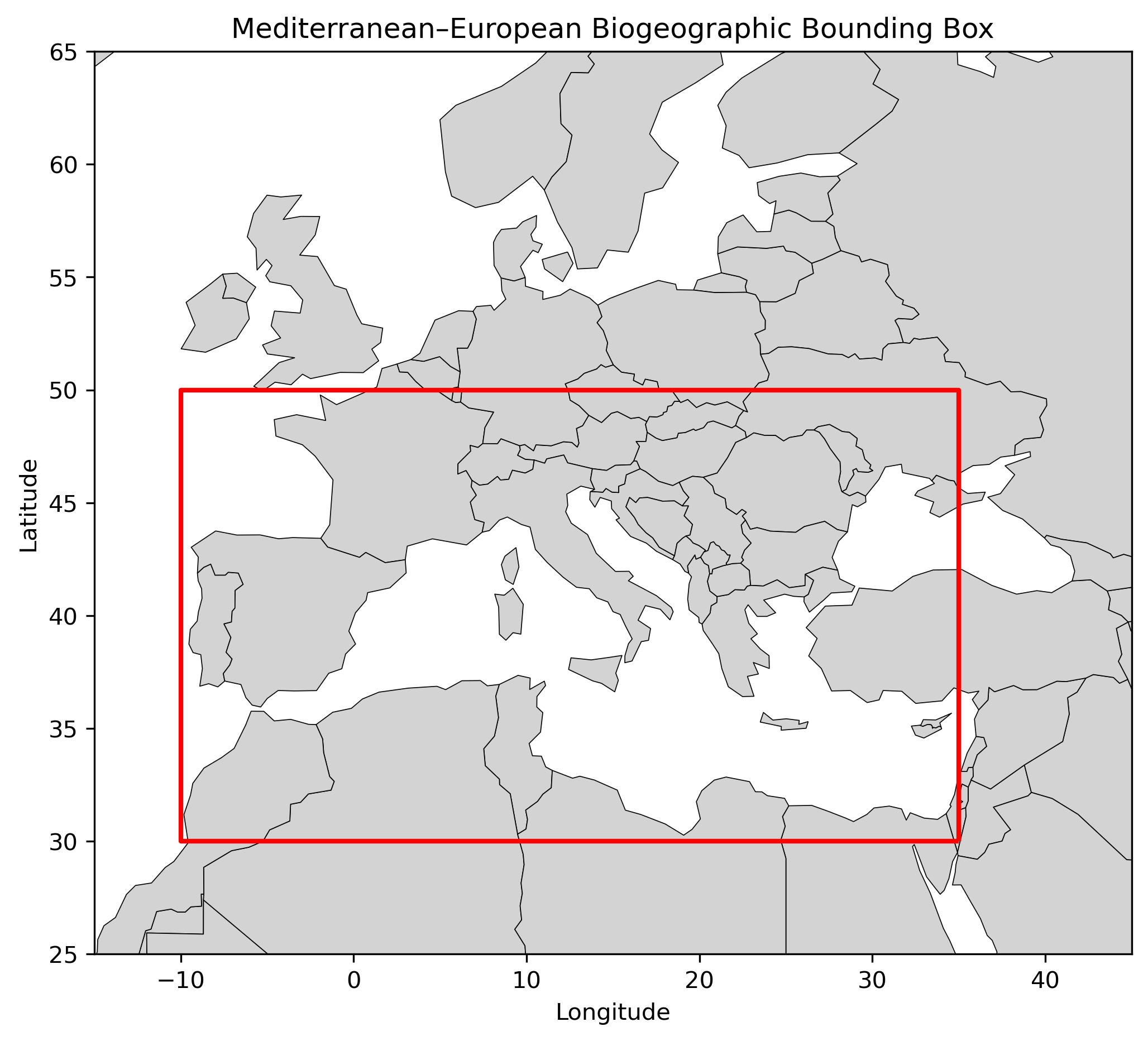}
  \caption{Geographical Distribution of plants explaining Voynich pictures (see table in Appendix (supplementary material page 12) section "Mapping Table (plant,scientific name)") }
  \label{fig:geospatialplant}
    \Description{Map showing the bounding box of the Mediterranean region indicating the geographic distribution of plants referenced in Voynich Manuscript illustrations, corresponding to the table in the supplementary material.}

\end{figure}

We designed a ChatGPT-based prompt to construct a geographical bounding box derived from the set of candidate taxa proposed to explain the Voynich illustrations (see Question 4 in Appendix (supplementary material page 11) section "ChatGPT prompting for Plant identification". As shown in Figure \ref{fig:geospatialplant}, the inferred distributions of these taxa are concentrated in Southern and Central Europe, North Africa, and Anatolia, encompassing the Mediterranean basin.

\subsection{Summary}

After conducting 874 queries using one large multimodal model (LMM), one small language model (SLM), and one hybrid-based image retrieval system (HBIR) , and examining the results corresponding to 23,720 image consultations, we derive the following conclusions:

\begin{assumption}
In Voynich MS the number of plant pictures (129) is very close to number of plant pictures in the famous Pseudo-Apulius (131).
\end{assumption}

\begin{assumption}
In Voynich MS pictures are inspired from medieval herbarius and notably Pseudo-Apulius but not only.
\end{assumption}

\begin{assumption}
In Voynich MS plants represented comes from Mediterranean area.
\end{assumption}

\section{Complementary pictures}
\label{sec:comppicture}

Between folio 57v (page 114) and folio 102v (page 204), drawings are present, though they appear individually rather than as part of continuous sequences.  
The placement of these illustrations may be guided by the accompanying recipes, while certain diagrams indicate the existence of an underlying cosmological or life cycle frameworks.\\
Some diagrams may have been inspired by Ramon Llull’s machines (early 14th century), which were widely known during the medieval period and referred to as the \emph{Ars Magna} (“Great Art”) \cite{LlullsGreatUniversalArt}.  
Llull devised and constructed a “logical machine,” likely influenced by the Arabic \emph{zairja}.
Theological theories, including subjects and predicates, were organized within geometric figures regarded as perfect, such as circles, squares, and triangles.  
By manipulating dials, levers, cranks, and rotating wheels, propositions and theses could be moved along designated guides, positioning themselves according to their assigned truth value—positive (true) or negative (false).
He also employed an \emph{astrolabium nocturnum}, which determined the passage of time based on the position of a star.  
Llull further holds the distinction of being the first in Europe to write on science and philosophy in languages other than Latin and Greek.
Similarly, the \emph{Ars Memoriae} (Art of Memory) represents a medieval mnemonic technique that utilizes images, spatial loci, and symbolic figures to organize reasoning and enhance memory \cite{Yates1966}.
The objects depicted within these mnemonic frameworks may be entirely imaginary or allegorical; their significance resides not in their material reality, but in their ability to support cognitive processes and the internalization of knowledge.\\
Both systems exemplify the interplay between symbolic representation and the organization of knowledge in specialized and medieval thought.

\section{Rare Event Detection: The Initial-Letter Sequence Hypothesis}
\label{sec:eventdet}

\subsection{Rare event problem definition}

When a sentence in a natural language is produced, it is possible to extract specific components in order to construct a derived artifact. 
We define an artifact obtained by taking the initial letter of each word, thereby forming what may be called a sentence fingerprint. For example, from the sentence “A boat breaks ice”, the resulting fingerprint is \emph{abbi}. 
By introducing additional modifiers—such as in “A blue British boat breaks ice”—the fingerprint becomes \emph{abbbbi}, exhibiting a repetition of the letter \emph{b}.\\
In this case, the sentence is grammatically correct but artificially constructed rather than extracted from an existing document. This raises a natural question: what is the probability of observing such repetitions of a given initial letter (here, four occurrences of b) within sentence fingerprints derived from a corpus of natural language documents?\\
More generally, this question concerns the statistical likelihood of repeated initials arising from the distribution of word-initial letters in a language, given typical sentence lengths and lexical frequencies.

Let $\mathcal{L}$ be a natural language with alphabet $\Sigma$, and let a sentence be modeled as a finite sequence of words
\[
S = (w_1, w_2, \dots, w_n),
\]
where each word $w_i$ is a string over $\Sigma$.

\begin{definition}
Given a sentence $S$, we define its \emph{sentence fingerprint} as the sequence
\[
F(S) = (f_1, f_2, \dots, f_n),
\]
where $f_i \in \Sigma$ denotes the initial letter of the word $w_i$.
\end{definition}

For example, the sentence \emph{``A boat breaks ice''} yields the fingerprint $\emph{abbi}$.  
By introducing additional modifiers, sentence length increases and repetitions may appear in the fingerprint. For instance, the grammatically correct sentence \emph{``A blue British boat breaks ice''} produces the fingerprint $\emph{abbbbi}$, which contains a run of four identical symbols.

We consider sentences drawn from a corpus $\mathcal{C}$ of natural language texts. Let $P$ denote the probability distribution over $\Sigma$ induced by the empirical frequencies of word-initial letters, and let $L$ be a probability distribution governing sentence length.

\begin{problem}[Rare initial-letter repetition]
Given a letter $x \in \Sigma$, an integer $k \geq 2$, and a sentence length $n$, estimate the probability that the sentence fingerprint $F(S)$ contains a contiguous subsequence of length $k$ composed entirely of the letter $x$.
\end{problem}

Equivalently, we seek to evaluate
\[
\mathbb{P}\bigl(\exists\, i \in \{1,\dots,n-k+1\} \text{ such that }
f_i = f_{i+1} = \dots = f_{i+k-1} = x \bigr),
\]
under the joint distribution induced by $P$ and $L$.

More generally, the objective is to characterize the statistical likelihood of long runs of identical word-initial letters in sentence fingerprints derived from natural language corpora, and to determine whether such patterns should be regarded as rare events or as expected outcomes given empirical letter frequencies and typical sentence lengths.

\subsection{Empirical observations}

\begin{figure}[htbp]
    \centering
    \includegraphics[width=1.1\linewidth]{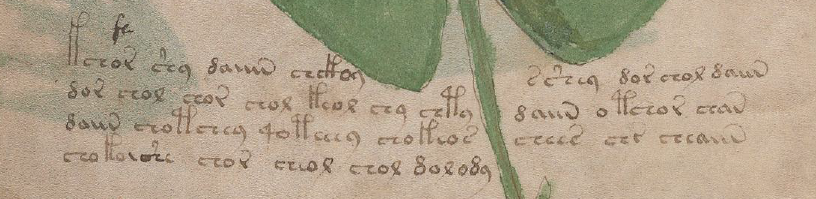}
    \caption{Illustration of rare events with repeated initials in in VM (folio 2v). This feature can be seen in the final two lines of the image.}
    \label{fig:rare-initials}
        \Description{Image from folio 2v of the Voynich Manuscript showing rare occurrences of repeated initials, particularly visible in the last two lines of the text.}

\end{figure}

As shown in Figure~\ref{fig:rare-initials} (Voynich MS, folio 2v, line~\#8), the glyph K1 appears eight times consecutively as a word-initial symbol. Table~\ref{tab:List9letterseq} lists the three glyph sequences in which such consecutive repetitions of word-initial symbols occur.

\begin{table}[htbp]
\centering
\begin{tabular}{|c|c|}
\hline
\textbf{Folio} & \textbf{Voynich MS sequences} \\
\hline
fRos.62 & \makecell[l]{A1C1.A1C2A1A3.A1Q2M1B1A2.A1Q1A3B2.A1Q1A1B1A3C1.\\
.A1Q2L1J1.A1Q1A1B1A2.A1Q2J1B1A3.} \\
\hline
f2v & \makecell[l]{K1A1Q1J1A1C2.K1K2C2.K1C1.K1J1A3G1.\\
K1A1Q1A1E1L1J1.K1A1C1.K1J1A1B2.K1A1B2.} \\
\hline
f8v & \makecell[l]{K1A1B2.K1A3E2.K1S1A2.K1A1C1.\\
K1J1A3F1.K1A3C1.K1K2Q1A2.K1A1C1.} \\
\hline
\end{tabular}
\caption{Sequences of 8 word-initial sequences with K1 and A1 symbols.}
\label{tab:List9letterseq}
\end{table}

\subsection{Model (Assumptions)}

Consider a sentence consisting of \( k \) words (more generally, a text may comprise multiple sentences, to which the same principle applies).
We examine all consecutive windows of eight words; when \( k \geq 8 \), this yields \( k - 7 \) such windows.

\textbf{Simplifying baseline assumption:} the initial letter of each word is independent, and each letter of the alphabet (47 letters) is equally likely.  
The effect of a non-uniform distribution will be discussed later.

\subsection{Probability that a given window of 8 words starts with the same letter}

Under the assumptions of uniformity and independence, the probability that 8 consecutive words begin with the same letter is

\[
p = 26 \times \left(\frac{1}{26}\right)^8=26 ^{-7}.
\]

Numerically, this evaluates to

\[
p \approx 1.245 \times 10^{-10}.
\]

\subsection{Probability that at least one such sequence occurs in a sentence of k words}

There are $k-7$ possible consecutive windows of 8 words.  
Assuming approximate independence between windows (a reasonable approximation for an order-of-magnitude estimate), the probability that at least one window consists of 8 words starting with the same letter is given by

\[
P(k) = 1 - (1 - p)^{k-7}.
\]

\subsection{Numerical examples (uniform model)}

Using the uniform model described above, we obtain the following values:

\begin{table}[ht]
\centering
\begin{tabular}{cc}
\hline
Number of words $k$ & Probability $P(k)$ \\
\hline
$8$      & $1.245 \times 10^{-10}$ \\
$20$     & $1.62 \times 10^{-9}$   \\
$100$    & $1.16 \times 10^{-8}$   \\
$1\,000$ & $1.24 \times 10^{-7}$   \\
$10\,000$ & $1.24 \times 10^{-6}$  \\
$50\,000$ & $6.2 \times 10^{-6}$  \\
\hline
\end{tabular}
\caption{Probability of observing at least one sequence of 8 consecutive words starting with the same letter under the uniform model.}
\label{tab:probabilities_uniform}
\end{table}

Even for a text containing tens of thousands of words, the probability remains extremely small under this model.

\subsection{Effect of a non-uniform distribution (more realistic for French)}

In French, the initial letters of words are not uniformly distributed: certain letters (such as \emph{s}, \emph{c}, \emph{l}, \emph{d}, \emph{p}, etc.) occur much more frequently at the beginning of words.

The true probability that a given window of 8 words all start with the same letter is therefore
\[
p_{\text{real}} = \sum_{a \in \text{letters}} f_a^8,
\]
where $f_a$ denotes the relative frequency of letter $a$ as an initial letter.

For a ``common'' letter with $f \approx 0.12$, the corresponding contribution is
\[
f^8 \approx 4.3 \times 10^{-8}.
\]

A small number of frequent letters can thus increase $p$ by several orders of magnitude compared to the uniform model (which yields $p \approx 1.25 \times 10^{-10}$).  
However, the resulting probabilities remain very small (typically in the range $10^{-8}$–$10^{-6}$ for a single window, depending on the precise distribution).

In other words, non-uniformity increases the probability, but not enough to make the event common in short sentences.

\subsection{Minimum corpus size to find a sequence: assumptions}

We consider a corpus of $40{,}000$ words, yielding $40{,}000 - 7 = 39{,}993$ consecutive windows of 8 words.

We assume that the initial letter of each word is approximately independent and follows a realistic distribution of initial letters in French (we use a plausible normalized vector of initial-letter frequencies, ignoring accents and ligatures).

We aim to compute:
\begin{itemize}
    \item the expected number of sequences of 8 consecutive words all starting with the same letter;
    \item the probability of observing at least one such sequence in the corpus.
\end{itemize}

\subsection{Expected number of sequences}

Let $f_a$ denote the relative frequency of letter $a$ as a word-initial letter.  
The probability that a given window of 8 words all start with the same letter is

\[
p_{\text{real}} = \sum_{a} f_a^8.
\]

The expected number of such sequences in the corpus is therefore

\[
\mathbb{E}[N] = (40{,}000 - 7)\, p_{\text{real}}.
\]

\subsection{Probability of observing at least one sequence}

Assuming approximate independence between windows, the probability of observing at least one such sequence is

\[
P = 1 - (1 - p_{\text{real}})^{39{,}993}.
\]

For a given letter $a$ with initial-letter frequency $f_a$, the probability that a window of 8 consecutive words all begin with the letter $a$ is $f_a^8$.

Summing over all letters, the probability that a given window of 8 words consists entirely of words starting with the same letter is
\[
p = \sum_a f_a^8.
\]

Over the entire corpus, the expected number of such sequences is
\[
\mathbb{E}[N] = (40{,}000 - 7)\, p.
\]

Assuming approximate independence between windows, the probability of observing at least one such sequence can be approximated using a Poisson model (which is valid here since the expected value is very small):
\[
\mathbb{P}(N \ge 1) \approx 1 - \exp\bigl(-\mathbb{E}[N]\bigr).
\]

\subsection{Numerical results}

Using a realistic distribution of initial letters in French, we obtain
\[
p = \sum_a f_a^8 \approx 1.50 \times 10^{-8}.
\]

The expected number of sequences of 8 consecutive words in a corpus of 40{,}000 words is therefore
\[
\mathbb{E}[N] \approx 0.000601.
\]

In other words, on average one expects only $0.000601$ such sequences in a corpus of 40{,}000 words—far below one.

The probability of observing at least one such sequence in a corpus of this size is
\[
\mathbb{P}(N \ge 1) \approx 0.0006006,
\]
that is, approximately $0.06006\%$ (about 6 chances in 10{,}000).

An alternative way to interpret this result is that one would need, on average, about $66.6$ million windows (roughly \textbf{\textit{66.6 million words}} ) to obtain an expected single occurrence.  
Thus, such an event is extraordinarily rare in ordinary text.

\subsection{Shorter identical initial-letter sequences}

\begin{table}[ht]
\centering
\begin{tabular}{c c c c c}
\hline
Length $L$ (per window) & $p_L$ & Windows & Expected value $E$ & Probability $\ge 1$ occurrence \\
\hline
5 words & $3.38 \times 10^{-5}$ & 39,996 & 1.353 & 0.741 (74.1\%) \\
6 words & $2.58 \times 10^{-6}$ & 39,995 & 0.103 & 0.098 (9.8\%) \\
7 words & $1.97 \times 10^{-7}$ & 39,994 & 0.00787 & 0.00784 (0.784\%) \\
8 words & $1.50 \times 10^{-8}$ & 39,993 & 0.000600 & 0.000600 (0.0600\%) \\
\hline
\end{tabular}
\caption{Probability of observing sequences of $L$ consecutive words starting with the same letter in a corpus of 40,000 words.}
\label{tab:sequence_probabilities}
\end{table}

For sequences of 5 consecutive words all starting with the same letter, the event is not rare: on average, we expect approximately 1.35 occurrences in a 40,000-word corpus (probability $\approx 74\%$ of observing at least one).\\

For sequences of 6 words, the event becomes already uncommon (expected value $\approx 0.10$ → $\approx 9.8\%$ chance).\\

For sequences of 7 or 8 words, the event is extremely rare in a corpus of this size (probabilities $\ll 1\%$).\\

\subsection{Summary}

\begin{assumption}
The probability that a sentence contains eight consecutive words beginning with the same initial letter is negligible in natural language corpora, except in extremely large datasets (on the order of millions of words) or in artificially constructed texts.
\end{assumption}

\section{Probabilistic model for word generation}

In this section, we model the generation of a Voynich Manuscript (VM) word sequence by assuming that each letter within a word is generated independently through a random sampling process governed by a given probability distribution.\\

\label{sec:WordGen}

Let $\mathcal{L} = \{A_1, A_2, A_3, B_1, K_1, \dots, V_1\}$ denote the finite set of available letters, with $|\mathcal{L}| = 47$.

We define subsets of \emph{high-probability letters} for different positions in a word:
\[
\mathcal{H} = \{A_1, A_2, B_1, G_1, K_1\}, \quad 
\mathcal{P} = \{A_1, B_2, K_1, J_1\}, \quad
\mathcal{S} = \{A_2, A_3, B_1, B_2, Q_1\},
\]
where $\mathcal{P}$ and $\mathcal{S}$ correspond to letters with elevated probability for prefixes and suffixes, respectively.

\subsection{Word length distribution}

Let $L \in \{3,4,5\}$ denote the random length of a word. We assume a uniform distribution:
\[
\Pr(L = \ell) = \frac{1}{3}, \quad \ell \in \{3,4,5\}.
\]

\subsection{Letter selection}

A word $W$ of length $L$ is represented as a sequence of letters:
\[
W = (w_1, w_2, \dots, w_L), \quad w_i \in \mathcal{L}.
\]

The letters are drawn independently conditional on position, but with the \emph{no-consecutive-repetition constraint}:
\[
\Pr(w_i = l \mid w_1, \dots, w_{i-1}) = 0 \quad \text{if } l = w_{i-1}.
\]

Define position-dependent probability mass functions (PMFs):

\begin{itemize}
  \item \textbf{Prefix letters ($i = 1,2,3$):}
  \[
  \Pr(w_i = l \mid w_{i-1}) = 
  \frac{w_{\mathrm{prefix}}(l)}{\sum_{l' \in \mathcal{L} \setminus \{w_{i-1}\}} w_{\mathrm{prefix}}(l')}, 
  \quad l \neq w_{i-1},
  \]
  where
  \[
  w_{\mathrm{prefix}}(l) =
  \begin{cases}
  20, & l \in \mathcal{P},\\
  1, & l \notin \mathcal{P}.
  \end{cases}
  \]

  \item \textbf{Middle letters ($i = 4, \dots, L-3$ if $L>5$):}
  \[
  \Pr(w_i = l \mid w_{i-1}) = 
  \frac{w_{\mathrm{middle}}(l)}{\sum_{l' \in \mathcal{L} \setminus \{w_{i-1}\}} w_{\mathrm{middle}}(l')}, 
  \quad l \neq w_{i-1},
  \]
  with
  \[
  w_{\mathrm{middle}}(l) =
  \begin{cases}
  10, & l \in \mathcal{H},\\
  1, & l \notin \mathcal{H}.
  \end{cases}
  \]

  \item \textbf{Suffix letters ($i = L-2, L-1, L$):}
  \[
  \Pr(w_i = l \mid w_{i-1}) = 
  \frac{w_{\mathrm{suffix}}(l)}{\sum_{l' \in \mathcal{L} \setminus \{w_{i-1}\}} w_{\mathrm{suffix}}(l')}, 
  \quad l \neq w_{i-1},
  \]
  where
  \[
  w_{\mathrm{suffix}}(l) =
  \begin{cases}
  20, & l \in \mathcal{S},\\
  1, & l \notin \mathcal{S}.
  \end{cases}
  \]
\end{itemize}

\subsection{Word generation process}

Words $W^{(1)}, \dots, W^{(N)}$ are generated independently according to the process described above:

\[
W^{(k)} = (w_1^{(k)}, \dots, w_{L^{(k)}}^{(k)}), \quad k = 1, \dots, N.
\]

The empirical frequency of a word $w$ in the generated corpus is

\[
f(w) = \sum_{k=1}^N \mathbf{1}_{\{W^{(k)} = w\}}.
\]

Sorting the words by $f(w)$ produces the empirical distribution of generated words.

\subsection{Constraints and notes}

These are rule for sampling :

\begin{enumerate}
  \item No consecutive letter repeats: $w_i \neq w_{i-1}$ for all $i \ge 2$.
  \item Prefix and suffix letters are given higher selection probability, as defined by $w_{\mathrm{prefix}}$ and $w_{\mathrm{suffix}}$.
  \item Middle letters favor the high-probability set $\mathcal{H}$.
\end{enumerate}

This model defines a \textit{position-dependent weighted categorical distribution} with memory of length 1 (to avoid repeats) and produces words of random length 3–5.

\subsection{Parameters Discussion}

These are the parameters of the distribution. Appendix (supplementary material page 46) section "Generative algorithm for word generation" presents the corresponding algorithm.

\begin{enumerate}
    \item \textbf{Consecutive-letter constraint:} The rule $w_i \neq w_{i-1}$ prevents repeated letters in a word.
    \item \textbf{Number of generated words ($N$):} Total words in the simulated corpus.
    \item \textbf{Word length distribution:} Random word length $L$ sampled from a specified range.
    \item \textbf{Weight values:} 
    \begin{itemize}
        \item \texttt{weights\_prefix} (default 20 for prefix letters, 1 otherwise)
        \item \texttt{weights\_suffix} (default 20 for suffix letters, 1 otherwise)
        \item \texttt{weights\_middle} (default 10 for high-probability letters, 1 otherwise)
    \end{itemize}
    \item \textbf{High-probability letters:} 
    \begin{itemize}
        \item \texttt{high\_prob\_letters} (affects middle letters)
        \item \texttt{prefix\_letters} (affects the first few letters)
        \item \texttt{suffix\_letters} (affects the last few letters)
    \end{itemize}

In our model, letters with high occurrence probability are designated as prefix or suffix letters. Specifically, the prefix letters are {A1,B2,K1,J1}, while the suffix letters are {A2,A3,B1,B2,Q1}. The corresponding weights for prefix and suffix letters are set to 20, whereas the weight for all other letters (middle positions) is set to 10. Word lengths L are sampled randomly from the integers 3 to 5, and a total of 
N=50,000 words are generated.

\begin{figure}[htbp]
  \centering
  \includegraphics[width=0.9\linewidth]{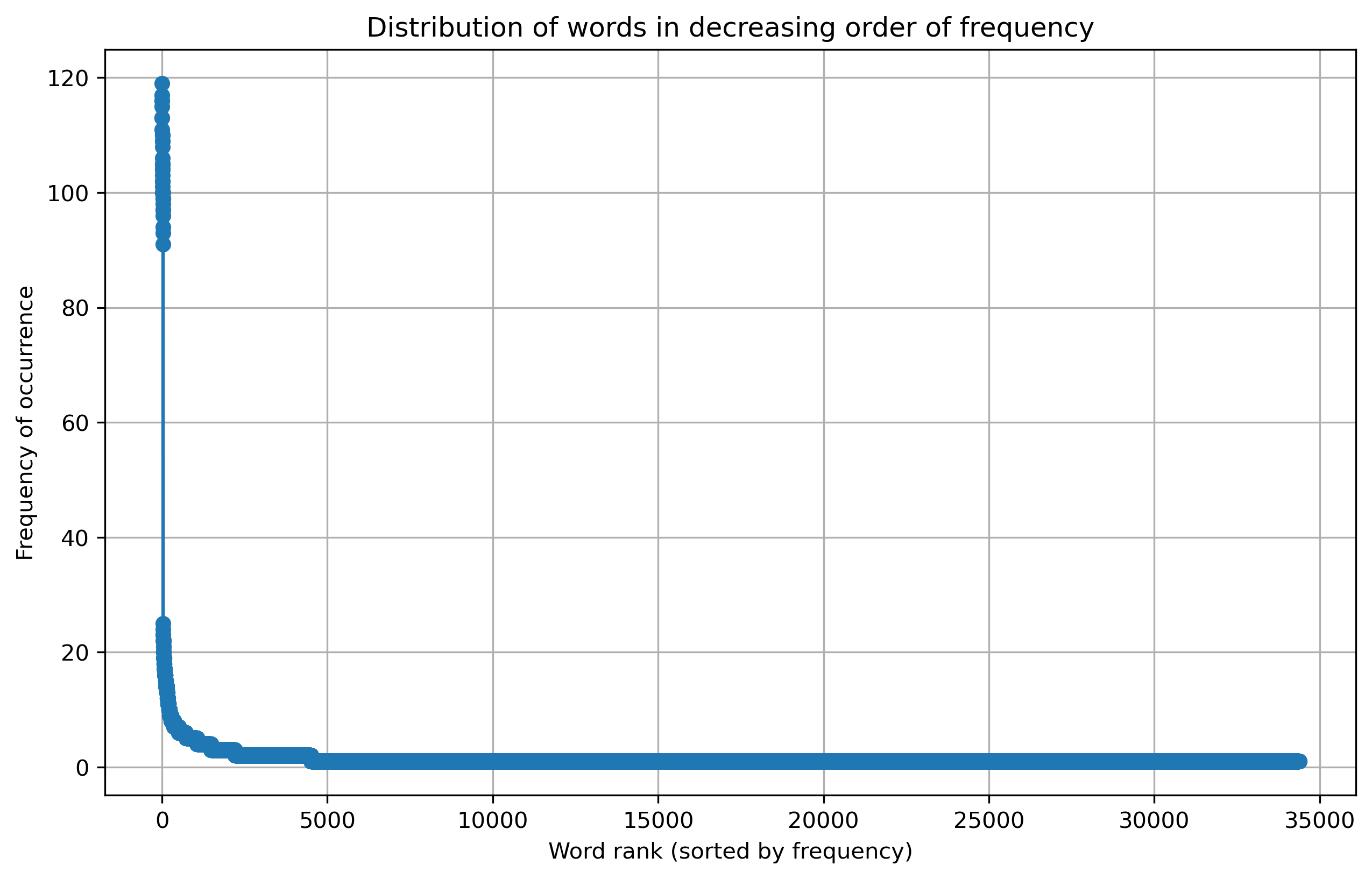}
  \caption{Distribution of words by decreasing frequency rank.}
  \label{fig:word-frequency}
    \Description{Bar chart showing the distribution of words in the Voynich Manuscript, ranked from most frequent to least frequent, illustrating the frequency drop-off pattern.}

\end{figure}

As shown in Figure~\ref{fig:word-frequency}, the word frequencies decrease rapidly with rank.

\end{enumerate}

\subsection{Summary}

\begin{assumption}
In the Voynich Manuscript, a probabilistic generative process with preferences for prefixes and suffixes can produce a Zipfian power-law distribution of word frequencies.
\end{assumption}

\subsection{Summary}

\begin{assumption}
The portion of the VM that lacks individual plant illustrations appears to be inspired by a logical system for deriving truth through geometric forms and structured reasoning, grounded in the integration of theology and philosophy, and closely related to the \emph{Ars Memoriae} and \emph{Ars Magna} systems.

\end{assumption}

\section{Conclusion}

This study situates the Voynich Manuscript (VM) within the long historical continuum of medical, botanical, and pharmacological knowledge transmission, drawing parallels with ancient and medieval traditions. 
Through a multidisciplinary approach that combines historical analysis, statistical modeling, and visual comparison, the manuscript emerges not as an isolated anomaly but as part of a broader tradition of technical and symbolic knowledge representation.
The predominance of plant imagery, the structured organization of the text, and the recurrence of statistical regularities argue against a purely random generation process and instead suggest an intentional design governed by internal constraints.\\
The Voynich Manuscript is a mysterious book whose content and authorship remain unclear. It comprises three main sections: the first consists of botanical illustrations accompanied by text resembling a medicinal herbal treatise (approximately 54\% of the manuscript); the second includes diagrams of cycles and recipe-like texts; and the final section consists solely of textual paragraphs (approximately 12\%).\\
The Voynich Manuscript has long fascinated scholars and the public alike, giving rise to numerous hypotheses regarding its origin, including proposals that it is a hoax, an encoded form of Latin, Hebrew, or Arabic, or even a text originating from the Americas based on interpretations of certain plant illustrations as tropical species. Our hypothesis is grounded in a multimodal text–image analysis that combines a detailed examination of 129 plant drawings with cross-linguistic comparison and the identification of anomalous structural patterns.\\ \\

\begin{quote}
We propose that the Voynich Manuscript was produced as an imitation inspired 
by contemporary knowledge, practices, and popular books of its 
time—that is, as a form of \textbf{\emph{pastiche}}. This hypothesis differs
from interpretations that regard the manuscript as a hoax or as
meaningless gibberish. A \textbf{\emph{pastiche}} is functional in the sense that
it may serve artistic, or pedagogical purposes, even if it does
not encode a conventional linguistic message. 
\end{quote}

We used the digitized corpus of the fifteenth-century VM based on the Zandbergen transcription, comprising 232 manuscript pages. The corpus contains 423 distinct glyphs, of which only 47 account for 99.6\% of all occurrences. The number of potentially significant glyph bigrams and trigrams is limited to 69 and 365, respectively. Word frequencies follow a Zipfian distribution, which is a characteristic feature of natural language.
We interpret the novel glyphs as being inspired by the Cistercian numerical encoding system, and we relate the cyclical diagrams inspired by philosophical concepts found in the writings of Ramon Llull. Taken together, these elements support the hypothesis that the VM is an imitative construction rather than an arbitrary or purely stochastic creation.\\
A comparative analysis with medieval herbals and modern botanical representations reveals both similarities and divergences, particularly in the stylized depiction of plants, the selective emphasis on specific morphological features, and the apparent liberties taken with color, symmetry, and proportion. 
Following 874 queries conducted using one large multimodal model (LMM), one small multimodal language model (SLM), and one hybrid-based image retrieval system (HBIR), and the examination of 23,720 associated image consultations, we conclude that the plant drawings are primarily inspired by the \emph{Pseudo-Apuleius} tradition and depict species originating from the Mediterranean region. 
These findings further support the hypothesis that the Voynich Manuscript represents an imitative construction.\\
Converting the VM into a language common in medieval Western Europe, using the most frequent symbols, does not produce meaningful sequences of words.
In the VM, the distribution of symbol sequences exhibits letter-like behavior, whereas the distribution of word lengths is more consistent with that of syllables. 
These observations suggest that the manuscript is unlikely to represent a natural language text.\\
We observed several sequences in which the initial glyphs of eight consecutive words are identical. 
The probability of encountering eight consecutive words beginning with the same initial letter is negligible in natural language corpora, except in extremely large datasets (on the order of millions of words) or in artificially constructed texts. 
These observations argue against the VM representing a natural language text.\\
Finally, we show that the VM can exhibit certain features of a natural language text when words are generated using a model that incorporates the usage and preferential placement of specific glyph bigrams as prefixes and suffixes. 
This model is capable of reproducing the observed Zipfian distribution of word frequencies. 
However, these results do not support the hypothesis that the manuscript represents a genuine natural language text.\\
The use of unknown glyphs and the generation of glyph sequences according to simple combinatorial rules have made the decipherment of the Voynich Manuscript a challenging problem for decades. 
The manuscript is therefore best understood as a coherent artifact reflecting intellectual, cultural, and technical practices, embodying a synthesis of empirical observation, symbolic representation, and encoded communication characteristic of pre-modern knowledge systems.  
We suggest that the ability to create such an artificial text—one that has led a wide range of experts in data science, medieval history, and linguistics to perceive it as potentially readable—may provide insights into the structural properties of natural language and its representations. 
Importantly, the integration of quantitative approaches with historical scholarship illustrates the value of methodological pluralism in addressing long-standing and unresolved problems.

\bibliographystyle{ACM-Reference-Format}
\bibliography{sample}





\section*{Ethics Statement and Data Availability}

This study adheres to standard ethical guidelines for research. The datasets used and/or analysed during the current study available from the corresponding author on reasonable request.

\section*{Declaration of competing interest }

The corresponding author is responsible for submitting a \href{http://www.nature.com/srep/policies/index.html#competing}{competing interests statement}. The author declares that there is no conflict of interest regarding the publication of this study.



\section*{Supplementary materials}

Supplementary data supporting the findings and the statistical analysis of this study are provided as : Supplementary material (PDF), 51 pages

\end{document}